\documentclass[10pt,journal,compsoc]{IEEEtran}
\usepackage[nocompress]{cite}
\usepackage{booktabs} 
\usepackage[normalem]{ulem}
\usepackage{subcaption}
\usepackage{graphicx}
\usepackage{amsmath}
\usepackage{amssymb}
\usepackage{color}
\usepackage{algorithmic,algorithm}
\usepackage{hyperref}
\usepackage{multirow}
\newcommand{\argmax}{\mathop{\rm arg~max}\limits}

\usepackage{mathtools}
\usepackage{enumitem}
\newcommand{\defeq}{\vcentcolon=}

\newenvironment{tight_itemize}{
\begin{itemize}
  \setlength{\itemsep}{0pt}
  \setlength{\parskip}{0pt}
  \setlength{\topsep}{0pt}
  \setlength{\partopsep}{0pt}
}{\end{itemize}}

\newenvironment{tight_enumerate}{
\begin{enumerate}
  \setlength{\itemsep}{0pt}
  \setlength{\parskip}{0pt}
  \setlength{\topsep}{0pt}
  \setlength{\partopsep}{0pt}
}{\end{enumerate}}

\begin{document}

\title{Virtual Adversarial Training:\\
A Regularization Method for Supervised and Semi-Supervised Learning}

\author{Takeru Miyato$^{*,\dagger, \ddagger}$, Shin-ichi Maeda$^{*,\dagger}$, Masanori Koyama$^{\S,\dagger}$ and Shin Ishii$^{\dagger, \ddagger}$ %
\IEEEcompsocitemizethanks{
	\IEEEcompsocthanksitem Contact : \texttt{takeru.miyato@gmail.com}
    \IEEEcompsocthanksitem $^*$ Preferred Networks, Inc., Tokyo, Japan, $^\dagger$ Graduate school of Informatics, Kyoto University, Kyoto, Japan, $^\ddagger$ ATR Cognitive Mechanisms Laboratories, Kyoto, Japan, $^\S$ Department of Mathematical Science, Ritsumeikan University, Kyoto, Japan
}}


\IEEEtitleabstractindextext{%
\begin{abstract}
We propose a new regularization method based on virtual adversarial loss: a new measure of local smoothness of the conditional label distribution given input. 
Virtual adversarial loss is defined as the robustness of the conditional label distribution around each input data point against local perturbation. 
Unlike adversarial training, our method defines the adversarial direction without label information and is hence applicable to semi-supervised learning.
Because the directions in which we smooth the model are only ``virtually" adversarial, we call our method virtual adversarial training (VAT).
The computational cost of VAT is relatively low. 
For neural networks, the approximated gradient of virtual adversarial loss can be computed with no more than two pairs of forward- and back-propagations.
In our experiments, we applied VAT to supervised and semi-supervised learning tasks on multiple benchmark datasets. With a simple enhancement of the algorithm based on the entropy minimization principle, our VAT achieves state-of-the-art performance for semi-supervised learning tasks on SVHN and CIFAR-10. 
\end{abstract}

}

\maketitle
\IEEEpeerreviewmaketitle
\IEEEraisesectionheading{\section{Introduction}}
\IEEEPARstart{I}{N}
practical regression and classification problems, one must face two problems on opposite ends; underfitting and overfitting. 
On one end, poor design of model and optimization process can result in large error for both training and testing dataset~(underfitting). 
On the other end, the size of the sample that can be used to tune the parameters of model is always finite, and the evaluation of the objective function in practice will always be a mere empirical approximation of the true expectation of the target value over the sample space. Therefore, even with successful optimization and low error rate on the training dataset (training error), the true expected error (test error) can be large~\cite{akaike1998information, watanabe2009algebraic}~(overfitting).
The subject of our study is the latter.
Regularization is a process of introducing additional information in order to manage this inevitable gap between the training error and the test error.
In this study, we introduce a novel regularization method applicable to semi-supervised learning that identifies the direction in which the classifier's behavior is most sensitive. 

Regularization is often carried out by augmenting the loss function with a so-called regularization term, which prevents the model from overfitting to the loss function evaluated on a finite set of sample points.
From Bayesian standpoint, regularization term can be interpreted as a prior distribution that reflects our educated \textit{a priori} knowledge or belief regarding the model~\cite{Bishop:2006}.
A popular \textit{a priori} belief based on widely observed facts is that the outputs of most naturally occurring systems are smooth with respect to spatial and/or temporal inputs~\cite{wahba1990spline}. 
What often underlie this belief are the laws of physics governing the system of interest, which in many cases are described by smooth models based on differential equations~\cite{arnol2013mathematical}.
When we are constructing the probability model, this belief prompts us to prefer conditional output distribution $p(y|x)$ (or just \textit{output distribution} for short) that are smooth with respect to conditional input $x$. 

In fact, smoothing the output distribution often works to our advantage in actual practice.
For example, label propagation~\cite{zhu2002learning} is an algorithm that improves the performance of classifier by assigning class labels to unlabeled training samples based on the belief that close input data points tend to have similar class labels. 
Also, it is known that, for neural networks~(NNs), one can improve the generalization performance by applying random perturbations to each input in order to generate artificial input points and encouraging the model to assign similar outputs to the set of artificial inputs derived from the same point~\cite{bishop1995training}.
Several studies have also confirmed that this philosophy of \textit{making the predictor robust against random and local perturbation} is effective in semi-supervised learning. 
For selected examples, see \cite{wager2013dropout,bachman2014learning,rasmus2015semi, sajjadi2016regularization,laine2016temporal}.

However, \cite{szegedy2013intriguing} and \cite{goodfellow2014explaining} found a weakness in naive application of this philosophy. 
They found that standard isotropic smoothing via random noise and random data augmentation often leaves the predictor particularly vulnerable to a small perturbation in a specific direction, that is, the adversarial direction, which is the direction in the input space in which the label probability $p(y=k|x)$ of the model is most sensitive.
\cite{szegedy2013intriguing} and~\cite{goodfellow2014explaining}  experimentally verified that the predictors trained with the standard regularization technique such as $L_1$ and $L_2$ regularization are likely to make mistakes when the signal is perturbed in the adversarial direction, even when the norm of the perturbation is so small that it cannot be perceived by human eyes. 

Inspired by this finding, Goodfellow et al.~\cite{goodfellow2014explaining} developed \textit{adversarial training} that trains the model to assign to each input data a label that is similar to the labels to be assigned to its neighbors in the adversarial direction. 
This attempt succeeded in improving generalization performance and made the model robust against adversarial perturbation. 
Goodfellow et al.'s work suggests that the locally isotropic output distribution cannot be achieved by making the model robust against isotropic noise.
In retrospect, this observation is in fact quite intuitive.
If the distribution around a given input is anisotropic and the goal is to resolve this anisotropy, it does not make much sense to exert equal smoothing ``pressure'' into all directions.

Our proposed regularization technique is a method that trains the \textit{output distribution} to be isotropically smooth around each input data point by selectively smoothing the model in its most anisotropic direction.
In order to quantify this idea, we introduce a notion of virtual adversarial direction, which is a direction of the perturbation that can most greatly alter the output distribution in the sense of distributional divergence. Virtual adversarial direction is our very interpretation of the `most' anisotropic direction.
In contrast, adversarial direction introduced by Goodfellow et al. \cite{goodfellow2014explaining} at an input data point is a direction of the perturbation that can most reduce the model's probability of correct classification, or the direction that can most greatly ``deviate" the prediction of the model from the correct label.
Unlike adversarial direction, virtual adversarial direction can be defined on unlabeled data point, because it is the direction that can most greatly deviate the \textit{current inferred output distribution} from the status quo.
In other words, even in the absence of label information, virtual adversarial direction can be defined on an unlabeled data point as if there is a ``virtual'' label; hence the name ``virtual" adversarial direction.

With the definition of virtual adversarial direction, we can quantify the local anisotropy of the model at each input point without using the supervisory signal. 
We define the local distributional smoothness~(LDS) to be the divergence-based distributional robustness of the model against virtual adversarial direction.
We propose a novel training method that uses an efficient approximation in order to maximize the likelihood of the model while promoting the model's LDS on each training input data point. 
For brevity, we call this method \textit{virtual adversarial training} (VAT). 

The following list summarizes the advantages of this new method: 
\begin{tight_itemize}
\item applicability to semi-supervised learning tasks
\item applicability to any parametric models for which we can evaluate the gradient with respect to input and parameter
\item small number of hyperparameters
\item parametrization invariant regularization
\end{tight_itemize}
The second advantage is worth emphasizing.
At first glance, our algorithm may appear as if it needs to solve an internal optimization problem in order to determine the virtual adversarial direction. 
For models such as NNs for which we can evaluate the gradient of the output with respect to the input, however, virtual adversarial perturbation admits an approximation that can be computed efficiently with the power method~\cite{golub2000eigenvalue}. 
This property enables us to implement VAT for NNs with no more than three times the computational cost of the standard, regularization-free training. 
This approximation step is an important part of the VAT algorithm that makes it readily applicable for various settings and model architectures.

Finally, the fourth advantage is not to be overlooked, because this is the most essential point at which our VAT is fundamentally different from popular regularization methods like $L_p$ regularization. For linear models, $L_p$ regularization has an effect of mitigating the oversensitivity of the output with respect to input, and one can control the strength of its effect via the hyperparameters. 
When the model in concern is highly nonlinear, as in the case of neural networks, however, the user has little control over the effect of $L_p$ regularization. 
Manipulation of the parameters in the first layer would have different effect on the final output depending on the choice of the parameters in the middle layers, and the same argument applies to the effect of regularization. 
In the language of Bayesian statistics with which we interpret the regularization term as prior distribution, this is to say that the nature of the prior distributions favored by the $L_p$ regularization depends on the current parameter-setting and is hence ambiguous and difficult to assess. 
Parameterization invariant regularization, on the other hand, does not suffer from such a problem.
In more precise terms, by \textit{parametrization invariant regularization} we mean the regularization based on an objective function $L(\theta)$ with the property that the corresponding optimal distribution $p(X;\theta^*)$ is invariant under the  one-to-one transformation $\omega = T(\theta), \theta = T^{-1}(\omega)$. 
That is, $p(X;\theta^*) = p(X;\omega^*)$ where $\omega^* = \arg \min _{\omega}L(T^{-1}(\omega) ;D) $. 
VAT is a parameterization invariant regularization, because it directly regularizes the output distribution by its local sensitivity of the output with respect to input, which is, by definition, independent from the way to parametrize the model.

When we applied VAT to the supervised and semi-supervised learning for the permutation invariant task on the MNIST dataset, our method outperformed all contemporary methods other than some cutting-edge methods that use sophisticated network architecture.
We also applied our method to semi-supervised learning on CIFAR-10 and Street View House Numbers (SVHN) datasets, and confirmed that our method achieves superior or comparable performance in comparison to those of state-of-the-art methods.

This article also extends our earlier work \cite{miyato2015distributional} in five aspects:
\begin{tight_itemize}
\item clarification of the objective function
\item comparison between VAT and random perturbation training (RPT)\footnote{a downgraded version of VAT introduced in this paper that smooths the label distribution at each point with same force in all directions. Please see the detail definition of RPT in Section~\ref{subsec:vatvsrp}.}
\item additional set of extensive experiments
\item evaluation of the virtual adversarial examples
\item enhancement of the algorithm with entropy minimization
\end{tight_itemize}

\section{Related Works}
Many classic regularization methods for NNs regularize the models by applying random perturbations to input and hidden layers \cite{reed1992regularization, bishop1995training, srivastava2014dropout, Goodfellow-et-al-2016}.
An early work by Bishop  \cite{bishop1995training} showed that adding Gaussian perturbation to inputs during the training process is equivalent to adding an extra regularization term to the objective function.
For small perturbations, the regularization term induced by such perturbation behaves similarly to a class of Tikhonov regularizers \cite{tikhonov1977solutions}. 
The application of random perturbations to inputs has an effect of smoothing the input-output relation of the NNs.  
Another way to smooth the input-output relation is to impose constraints on the derivatives.
For example, constraints may be imposed on the Frobenius norm of the Jacobian matrix of the output with respect to the input.
This approach was taken by Gu and Rigazio~\cite{gu2014towards} in their deep contractive network. 
Instead of computing the computationally expensive full Jacobian, however, they approximated the Jacobian by the sum of the Frobenius norms of the layer-wise Jacobians computed for all adjacent pairs of hidden layers. 
Possibly because of their layer-wise approximation, however, deep contractive network was not successful in significantly decreasing the test error. 

Dropout \cite{srivastava2014dropout} is another popular method for regularizing NNs with random noise. Dropout is a method that assigns random masks on inputs/hidden layers in the network during its training. 
From a Bayesian perspective, dropout is a method to introduce prior distribution for the parameters\cite{maeda2014bayesian, gal2015dropout}. 
In this interpretation, dropout is a method that regularizes the model via Bayesian model ensembles, and is complementary to our approach, which directly augments the function with a regularization term.

Adversarial training was originally proposed by~\cite{szegedy2013intriguing}.
They discovered that some architectures of NNs, including state-of-the-art NN models, are particularly vulnerable to a perturbation applied to an input in the direction to which the models' label assignment to the input is most sensitive (adversarial), even when the perturbation is so small that human eyes cannot discern the difference. They also showed that training the models to be robust against adversarial perturbation is effective in reducing the test error. However, the definition of adversarial perturbation in \cite{szegedy2013intriguing} required a computationally expensive inner loop in order to evaluate the adversarial direction.
To overcome this problem, Goodfellow et al. \cite{goodfellow2014explaining} proposed another definition of adversarial perturbation that admits a form of approximation that is free of the expensive inner loop (see the next section for details).

Bachman et al.~\cite{bachman2014learning} studied the effect of random perturbation in the setting of semi-supervised learning.
Pseudo Ensemble Agreement introduced in~\cite{bachman2014learning} trains the model in a way that the the output from each layer in the NN does not \textit{vary} too much by the introduction of random perturbations.
On the other hand, ladder networks~\cite{rasmus2015semi} achieved high performance for semi-supervised learning tasks by making an effort so that one can reconstruct the original signal of the lower layer from the signal of the uppermost layer and the noise-perturbed outputs from the hidden layers. 

Our method is similar in philosophy to~\cite{bachman2014learning}, and we use the virtual adversarial perturbation for their noise process.
As we show later, in our experiments, this choice of perturbation was able to improve the generalization performance. 
Random image augmentation is a variant of random perturbation that simply augments the dataset with images perturbed by regular deformation.
For more theoretical overview of related methods of noise regularizations, Jacobian regularizations and their extensions like PEA\cite{bachman2014learning}, ladder networks\cite{rasmus2015semi} and VAT, we refer the readers to ~\cite{abbas2016understanding}.  

Several works \cite{laine2016temporal,sajjadi2016regularization} succeeded in using the random image augmentation to improve generalization performance for semi-supervised tasks of image classification. 
These methods can also be interpreted as types of techniques that smooth the model around input data points and extrapolate the labels of unlabeled examples. 
In the area of nonparametric studies, this type of method is referred to as label propagation \cite{zhu2002learning}. 

Another family of methods for the semi-supervised learning of NNs that are worth mentioning is the family based on sophisticated generative models.
The methods belonging to this family are different from those that we introduced above because they do not require an explicit definition of smoothness. 
Kingma et al.~\cite{kingma2014semi} applied a variational-autoencoder-based generative model to semi-supervised learning. 
This work was followed by several variants \cite{maaloe2016auxiliary}.
Generative adversarial networks~(GANs) proposed by~\cite{goodfellow2014generative} are a recently popular high-performance framework that can also be applied to semi-supervised learning~\cite{springenberg2015unsupervised, salimans2016improved}.
In practice, these methods often require careful tuning of many hyperparameters in the generative model, and are usually not easy to implement without high expertise in its optimization process. 

\section{Methods}
We begin this section with a set of notations. 
Let $x \in R^I$ and $y \in Q$ respectively denote an input vector and an output label, where $I$ is the input dimension and $Q$ is the space of all labels. 
Additionally, we denote the output distribution parameterized by $\theta$ as $p(y|x, \theta)$. 
We use $\hat{\theta}$ to denote the vector of the model parameters at a specific iteration step of the training process. 
We use $\mathcal{D}_{l} = \{x_l^{(n)}, y_l^{(n)} | n = 1,\dots,N_l\}$ to denote a labeled dataset, and $\mathcal{D}_{ul} = \{x_{ul}^{(m)} | m = 1,\dots,N_{ul}\}$ to denote an unlabeled dataset. We train the model $p(y|x, \theta)$ using $\mathcal{D}_{l}$ and $\mathcal{D}_{ul}$.

\subsection{Adversarial Training}
Our method is closely related to the adversarial training proposed by Goodfellow et al. \cite{goodfellow2014explaining}.  We therefore formulate  adversarial training before introducing our method. 
The loss function of adversarial training in \cite{goodfellow2014explaining} can be written as
\begin{eqnarray}
L_{\rm adv}(x_l,\theta) \defeq D\left[q(y|x_l), p(y|x_l+r_{\rm adv},\theta)\right] \label{eq:adv_loss}\\
{\rm where}\ r_{\rm adv} \defeq \argmax_{r;\|r\|\leq\epsilon} D\left[q(y|x_l), p(y|x_l + r, \theta)\right], \label{eq:adv_ptb}
\end{eqnarray}
where $D[p, p']$ is a non-negative function that measures the divergence  between two distributions $p$ and $p'$. For example, $D$ can be the cross entropy $D[p,p']=-\sum_i p_i \log p_i'$, where $p$ and $p'$ are vectors whose $i$-th coordinate represents the probability for the $i$-th class. 
The function $q(y|x_l)$ is the true distribution of the output label, which is unknown.
The goal with this loss function is to approximate the true distribution $q(y|x_l)$ by a parametric model $p(y|x_l, \theta)$ that is robust against adversarial attack to $x$.
In \cite{goodfellow2014explaining}, the function $q(y|x_l)$ was approximated by one hot vector $h(y;y_l)$, whose entries are all zero except for the index corresponding to the true label~(output) $y_l$. 
Likewise, for regression tasks, we can use the normal distribution centered at $y_l$ with constant variance, or the delta function with the atom at $y = y_l$.

Generally, we cannot obtain a closed form for the exact adversarial perturbation $r_{\rm adv}$.
However, we can approximate $r_{\rm adv}$ with a linear approximation of $D$ with respect to $r$ in Eq.\eqref{eq:adv_ptb}. 
When the norm is $L_2$, adversarial perturbation can be approximated by
\begin{eqnarray}
r_{\rm adv} &\approx&  \epsilon \frac{g}{\|g\|_2},\ {\rm where}\ g=\nabla_{x_l} D\left[h(y;y_l), p(y|x_l, \theta)\right]. \label{eq:approx_adv_ptb_L2}
\end{eqnarray} 
When the norm is $L_{\infty}$, adversarial perturbation can be approximated by
\begin{eqnarray}
r_{\rm adv} &\approx&  \epsilon {\rm sign}(g), \label{eq:approx_adv_ptb_max}
\end{eqnarray}
where $g$ is the same function that appeared in Eq.\eqref{eq:approx_adv_ptb_L2}.
\cite{goodfellow2014explaining} originally used \eqref{eq:approx_adv_ptb_max} for their adversarial training.
Note that, for NNs, the gradient $\nabla_{x_l} D\left[h(y;y_l), p(y|x_l, \theta)\right]$ can be efficiently computed by backpropagation.
By optimizing the loss function of the adversarial training in Eq.\eqref{eq:adv_loss} based on the adversarial perturbation defined by Eq.\eqref{eq:approx_adv_ptb_L2} (or \eqref{eq:approx_adv_ptb_max}), \cite{goodfellow2014explaining} and~\cite{miyato2015distributional} were able to train a model with better generalization performance than the model trained with random perturbations.

\subsection{\label{subsec:vat}Virtual Adversarial Training}
Adversarial training is a successful method that works for many supervised problems.
However, full label information is not available at all times. Let $x_*$ represent either $x_l$ or $x_{ul}$.  Our objective function is now given by
\begin{align}
	&D\left[q(y|x_*),p(y|x_*+r_{\rm qadv},\theta)\right] \nonumber \\
    &{\rm where}\ r_{\rm qadv} \defeq \argmax_{r;\|r\|\leq\epsilon} D\left[q(y|x_*), p(y|x_*+r, \theta)\right] \nonumber,
\end{align} 
Indeed, we have no direct information about $q(y|x_{ul})$. We therefore take the  strategy to replace $q(y|x)$ with its current  approximation, $p(y|x, \theta)$. This approximation is not necessarily naive, because $p(y|x, \theta)$ shall be close to $q(y|x)$ when the number of labeled training samples is large.
This is also the motivation behind our inclusion of the term ``virtual" in our work.
Literally,  we use ``virtual" labels that are probabilistically generated from $p(y|x, \theta)$ in place of labels that are unknown to the user, and compute adversarial direction based on the virtual labels.

Therefore, in this study, we use the \textit{current} estimate $p(y|x,\hat{\theta})$ in place of $q(y|x)$.
With this compromise, we arrive at our rendition of Eq.\eqref{eq:adv_ptb} given by 
\begin{eqnarray}
	{\rm LDS}(x_*, \theta) \defeq D\left[p(y|x_*,\hat{\theta}),p(y|x_*+r_{\rm vadv},\theta)\right] \label{eq:vadv_loss}\\
  	r_{\rm vadv} \defeq \argmax_{r;\|r\|_{2}\leq\epsilon} D\left[p(y|x_*,\hat{\theta}), p(y|x_*+r)\right],
\end{eqnarray}
which defines our {\it virtual adversarial perturbation}.
The loss ${\rm LDS}(x, \theta)$ can be considered as a negative measure of the local smoothness of the current model at each input data point $x$, and its reduction would make the model smooth at each data point. 
The regularization term we propose in this study is the average of ${\rm LDS}(x_*, \theta)$ over all input data points:
\begin{eqnarray}
	\mathcal{R}_{\rm vadv}(\mathcal{D}_l, \mathcal{D}_{ul}, \theta) \defeq \frac{1}{N_l + N_{ul}}\sum_{x_* \in \mathcal{D}_l, \mathcal{D}_{ul}} {\rm LDS}(x_*, \theta) \label{eq:reg_vadv}.
\end{eqnarray}
The full objective function is thus given by
\begin{align}
\ell(\mathcal{D}_l, \theta) +  \alpha \mathcal{R}_{\rm vadv}(\mathcal{D}_l, \mathcal{D}_{ul}, \theta), \label{eq:full}
\end{align} 
where $\ell(\mathcal{D}_l, \theta)$ is the negative log-likelihood for the labeled dataset. 
VAT is a training method with the regularizer $\mathcal{R}_{\rm
vadv}$.

\begin{figure*}[ht]
	\centering
	\includegraphics[width=0.85\textwidth]{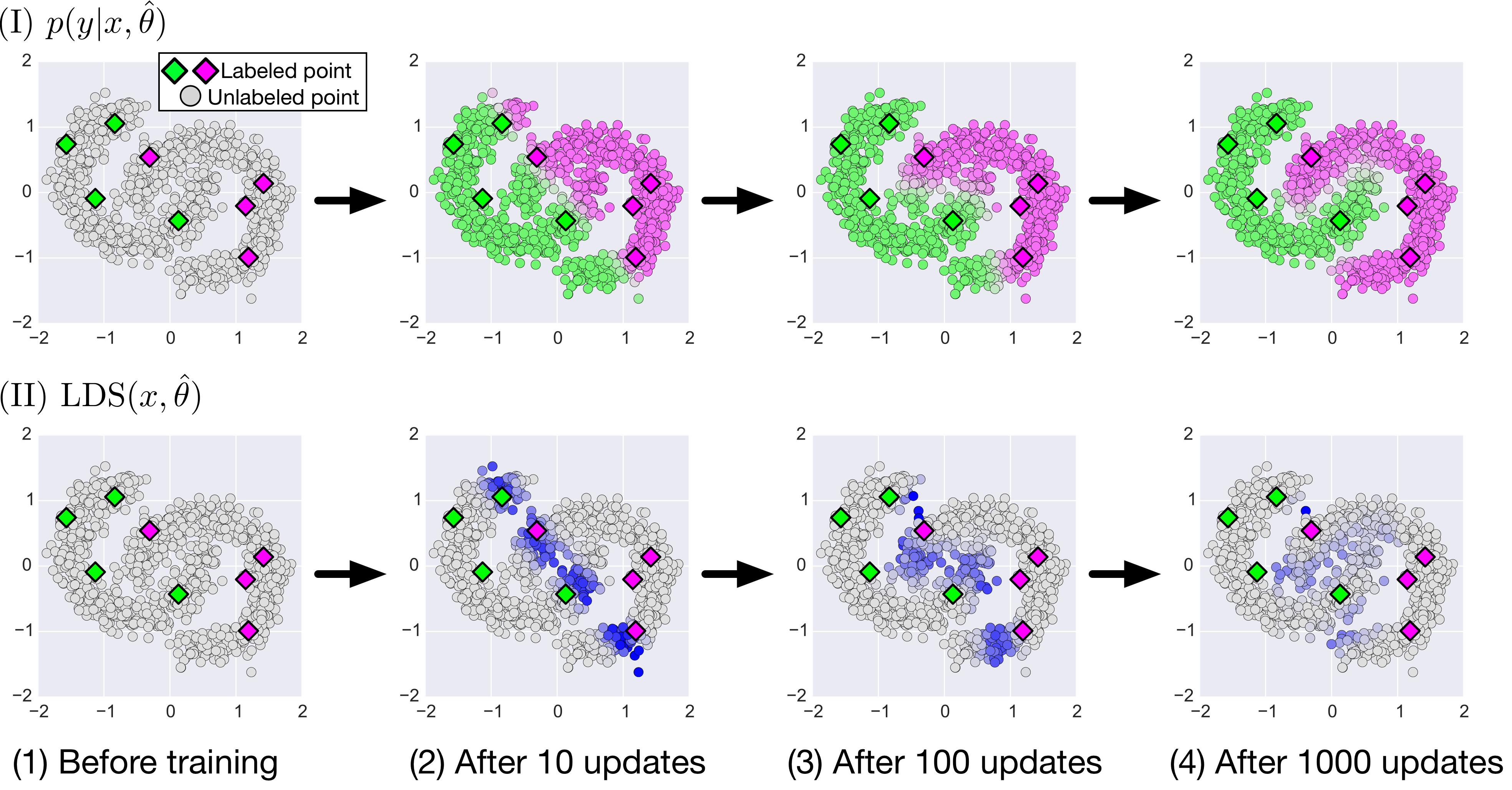}
	\caption{\label{fig:syn_cont} Demonstration of how our VAT works on semi-supervised learning. We generated 8 labeled data points ($y=1$ and $y=0$ are green and purple, respectively), and 1,000 unlabeled data points in 2-D space. 
    The panels in the first row (I) show the prediction $p(y=1|x,\theta)$ on the unlabeled input points at different stages of the algorithm. We used a continuous colormap to designate the predicted values of $p(y=1|x,\theta)$, with Green, gray, and purple respectively corresponding to the values 1.0, 0.5, and 0.0.
    The panels in the second row (II) are heat maps of the regularization term ${\rm LDS}(x,\hat{\theta})$ on the input points. 
	The values of LDS on blue-colored points are relatively high in comparison to the gray-colored points. 
    We used KL divergence for the choice of $D$ in Eq.\eqref{eq:vadv_loss}.
    Note that, at the onset of training, all the data points have similar influence on the classifier.
    After 10 updates, the model boundary was still appearing \textit{over} the inputs.
    As the training progressed, VAT pushed the boundary away from the labeled input data points.
}
\end{figure*}

One notable advantage of VAT is that there are just two scalar-valued hyperparameters: (1) the norm constraint $\epsilon>0$ for the adversarial direction and (2) the regularization coefficient  $\alpha>0$ that controls the relative balance between the negative log-likelihood and the regularizer $\mathcal{R}_{\rm vadv}$. 
In fact, for all our experiments, our VAT achieved superior performance by tuning only the hyperparameter $\epsilon$, while fixing $\alpha=1$.
Theoretically, these two hyperparameters play similar roles, as discussed later in Section~\ref{subsec:hp}.
One advantage of VAT is the number of hyperparameters. 
For many generative model-based supervised and semi-supervised learning methods aimed at learning $p(y,x)$, a bottleneck of training is the difficulty of the optimization of hyperparameters for the generative model (i.e. $p(x)$ or $p(x|y)$). 
Also, as opposed to adversarial training \cite{goodfellow2014explaining}, the definition of virtual adversarial perturbation only requires input $x$ and does not require label $y$.  
This is the property that allows us to apply VAT to semi-supervised learning. 
Fig. \ref{fig:syn_cont} shows how VAT works on semi-supervised learning on a two-dimensional synthetic dataset. 
We used an NN classifier with one hidden layer with 50 hidden units. 
At the beginning of the training, the classifier predicted different labels for input data points in the same cluster, and ${\rm LDS}$ on the boundaries were very high (see panel (2) in Fig. \ref{fig:syn_cont}).
The algorithm exerts high pressure for the model to be smooth around the points with large ${\rm LDS}$ values. 
As the training progressed, the model evolved so that the label prediction on the points with large ${\rm LDS}$ values are strongly influenced by the labeled inputs in the vicinity.  This encouraged the model to predict the same label for the set of points that belong to the same cluster, which is what we often desire in semi-supervised learning. 
As we can see in Fig.~\ref{fig:syn_cont}, VAT on semi-supervised learning can be given a similar interpretation as label propagation \cite{zhu2002learning}, which is another branch of method for semi-supervised learning. 



\subsection{\label{subsec:fastapprox}Fast Approximation Method for $r_{\rm vadv}$ 
and the Derivative of the Objective Function}
Once virtual adversarial perturbation $r_{\rm vadv}$ is computed, the evaluation of ${\rm LDS}(x_*, \theta)$ simply becomes the computation of the divergence $D$ between the output distributions $p(y |x_{*},\hat{\theta})$ and $p(y|x_{*} + r_{\rm vadv},\theta)$.
However, the evaluation of $r_{\rm vadv}$ cannot be performed with the linear approximation as in the original adversarial training (Eq.\eqref{eq:approx_adv_ptb_L2}) because the gradient of $D[p(y|x_{*},\hat{\theta}), p(y|x_{*}+r, \hat{\theta})]$ with respect to $r$ is always $0$ at $r=0$.
In the following, we propose an efficient computation of $r_{\rm vadv}$, for which there is no evident closed form. 

For simplicity, we denote $D[p(y|x_{*},\hat{\theta}), p(y|x_{*}+r, \theta)]$ by $D(r,x_{*}, \theta)$. 
We assume that $p(y |x_{*},\theta)$ is twice differentiable with respect to $\theta$ and $x$ almost everywhere. 
Because $D(r,x_*,\hat{\theta})$ takes the minimal value at $r = 0$, the differentiability assumption dictates that its first derivative $\nabla_r D(r,x,\hat{\theta})|_{r=0}$ is zero.  Therefore, the second-order Taylor approximation of $D$ is 
\begin{eqnarray}
	D(r,x,\hat{\theta}) &\approx&  \frac{1}{2}r^{T}H(x,\hat{\theta}) r, \label{eq:second_taylor}
\end{eqnarray}
where $ H(x,\hat{\theta})$ is the Hessian matrix given by $ H(x,\hat{\theta}) \defeq \nabla\nabla_{r} D(r,x,\hat{\theta})|_{r=0}$. 
Under this approximation, $r_{\rm vadv}$ emerges as the first dominant eigenvector $u(x,\hat{\theta})$ of $H(x,\hat{\theta})$ with magnitude $\epsilon$: 
\begin{eqnarray}
r_{\rm vadv} &\approx& \arg \mathop {\rm max}\limits_r \{r^{T} H(x, \hat{\theta}) r; ~\|r\|_2 \leq \epsilon\}  \nonumber \\
&=& \epsilon \overline{u(x, \hat{\theta})}, 
\end{eqnarray} 
where $\bar{v}$ denotes the unit vector whose direction is the same as its argument vector $v$; that is, $\bar{v} \equiv \frac{v}{\| v \|_2}$.
Hereafter, we denote $H(x,\hat{\theta})$ by $H$ for simplicity.

Next, we need to address the $O(I^3)$ runtime required for the computation of the eigenvectors of the Hessian $H$. We resolve this issue with the approximation via the power iteration method~\cite{golub2000eigenvalue} and the finite difference method.\\
Let $d$ be a randomly sampled unit vector. 
Provided that $d$ is not perpendicular to the dominant eigenvector $u$, the iterative calculation of
\begin{eqnarray}
 d \leftarrow \overline{H d} \label{eq:power_method}
\end{eqnarray}
makes $d$ converge to $u$. 
To reduce the computational time, we perform this operation without the direct computation of $H$. 
Note that $H d$ can be approximated using the finite difference method:
\begin{eqnarray}
	H d &\approx& \frac{\nabla_{r} D(r,x, \hat{\theta}) |_{r=\xi d} - \nabla_{r} D(r,x,\hat{\theta})|_{r=0}}{\xi} \nonumber \\
	&=& \frac{\nabla_{r}D(r,x,\hat{\theta})|_{r=\xi d}}{\xi}, \label{eq:dif_method}
\end{eqnarray}
with $\xi \neq 0$. 
In the computation above, we use the fact that $\nabla_r D(r,x,\hat{\theta})|_{r=0} = 0$ again. 
To summarize, we can approximate $r_{\rm vadv}$ with the repeated application of the following update: 
\begin{eqnarray}
d \leftarrow \overline{\nabla_{r} D(r,x,\hat{\theta})|_{r=\xi d}}.
\end{eqnarray}
The computation of $\nabla_r D$ can be performed in a straightforward manner. For NNs, this can be performed with one set of backpropagation. 
The approximation introduced here can be improved monotonically by increasing the number of the power iterations $K$. 
Thus, for NNs, the computation of $r_{\rm vadv}$ can be performed with $K$ sets of backpropagations. 
Surprisingly, only one power iteration was sufficient for high performance on various benchmark datasets. 
This approximation of $r_{\rm vadv}$ with $K=1$ results in an approximation that is similar in form to Eq.\eqref{eq:approx_adv_ptb_L2}:
\begin{align}
	&r_{\rm vadv} \approx  \epsilon \frac{g}{\|g\|_2} \\
    &{\rm where}\ g=\nabla_r D\left[p(y|x,\hat{\theta}), p(y|x+r, \hat{\theta})\right]\Big|_{r=\xi d}.
\end{align}
We further discuss the effects of the number of the power iterations in Section \ref{sec:experiments}.
After computing $r_{\rm vadv}$, the derivative of $\mathcal{R}_{\rm vadv}$ can be easily computed with one set of forward- and back-propagation on NNs.
Meanwhile, the derivative of $r_{\rm vadv}$ with respect to $\theta$ is not only convoluted and computationally costly, but also introduces another source of variance to the gradient and negatively affect the performance of the algorithm.
Our VAT therefore ignores the dependency of $r_{\rm vadv}$ on $\theta$. 
In total, the derivative of the full  objective function including the log-likelihood term \eqref{eq:full} can be computed with $K + 2$ sets of backpropagation. 
Algorithm \ref{algo:vat_loss} summarizes the procedure for mini-batch SGD with the approximation of $\nabla_{\theta} \mathcal{R}_{\rm vadv}$ carried out with one power iteration. 
VAT is an algorithm that updates the model by the weighted sum of the gradient of the likelihood and the gradient $\nabla_{\theta} \mathcal{R}_{\rm vadv}$ computed with Algorithm \ref{algo:vat_loss}.

\begin{algorithm}
\caption{\label{algo:vat_loss}Mini-batch SGD for $\nabla_{\theta} \mathcal{R}_{\rm vadv}(\theta)|_{\theta=\hat{\theta}}$, with a one-time power iteration method.}
\begin{tight_enumerate}
\item Choose $M$ samples of $x^{(i)} (i=1,\dots,M)$ from dataset $\mathcal{D}$ at random.
\item Generate a random unit vector $d^{(i)} \in R^{I}$ using an iid Gaussian distribution.
\item Calculate $r_{\rm vadv}$ via taking the gradient of $D$ with respect to $r$ on $r=\xi d^{(i)}$ on each input data point $x^{(i)}$:
\begin{tight_enumerate}
\item[] $g^{(i)} \leftarrow \nabla_{r} D\left[p(y|x^{(i)}, \hat{\theta}), p(y|x^{(i)}+r, \hat{\theta})\right]\Big|_{r=\xi d^{(i)}}$, 
\item[] $r_{\rm vadv}^{(i)} \leftarrow g^{(i)}/\|g^{(i)}\|_2$
\end{tight_enumerate}
\item \textbf{Return} 
\begin{eqnarray}
\nabla_{\theta} \left(\frac{1}{M}\sum_{i=1}^M D\left[p(y|x^{(i)}, \hat{\theta}), p(y|x^{(i)}+r_{{\rm vadv}}^{(i)}, \theta)\right]\right)\Bigg|_{\theta=\hat{\theta}} \nonumber \label{eq:approx_vat_loss}
\end{eqnarray}
\end{tight_enumerate}
\end{algorithm}

 \subsection{\label{subsec:vatvsrp}Virtual Adversarial Training vs. Random Perturbation Training}
The regularization function we use for VAT can be generally written as 
\begin{align}
&\mathcal{R}^{(K)}(\theta, \mathcal{D}_l, \mathcal{D}_{ul}) \nonumber \\
&\defeq \frac{1}{N_l + N_{ul}}\sum_{x\in \mathcal{D}_l, \mathcal{D}_{ul}} E_{r_K} \left[ D\left[p(y|x,\hat{\theta}),p(y|x+r_K,\theta)\right]\right], \label{eq:R_K}
\end{align}
where $r_K$ is obtained by applying the power iteration $K$-times on a sample from the uniform distribution on the sphere $U(r|\epsilon)$ with radius $\epsilon$.
In practice, for the computation of \eqref{eq:R_K} we use an empirical expectation about the random perturbation $r_K$.  
For the implementation of VAT, we use this regularizer with $K \geq 1$.
Meanwhile, We refer to the training with $\mathcal{R}^{(0)}$ as Random Perturbation Training~(RPT). 
RPT is a downgraded version of VAT that does not perform the power iteration. 
By definition, RPT only smooths the function isotropically around each input data point. 

As we discuss further in Section~\ref{sec:experiments}, RPT falls behind VAT in its sheer ability to reduce the generalization error. 
There could be two reasons for the superiority of VAT. 

First, the learning process of VAT is inherently more stable than that of RPT. 
At each step of the algorithm, the power iteration generates a vector that has a large projection to the virtual adversarial direction with high probability.
Note that, as $K \to \infty$, under the sufficient regularity of the model, the gradient of $D$ in the expression~\eqref{eq:R_K} approaches the deterministic vector $1/2\epsilon^2\nabla_\theta \lambda_1 (x, \theta)$, where $\lambda_1$ is the dominant eigenvalue of $H(x, \theta)$.

Thus, the direction to which VAT smooths the model is more deterministic than the direction to which RPT smooths the model, which is uniformly distributed over the sphere of radius $\epsilon$; the stability of the learning of RPT always suffers from the variance of \eqref{eq:R_K}. 

Second, the regularization function of RPT has an  essentially effect on the model.  
For each observed input point, VAT trains the model to assign similar label distribution only to the set of proximal points aligned in the virtual adversarial direction.
In contrast, RPT encourages the model to assign the same label distribution to all input points in the isotropic neighborhood of each observed input. 
From spectral perspective, the difference between VAT and RPT is that VAT penalizes the spectral norm~(largest singular value) of the Hessian matrix $H$~\eqref{eq:second_taylor}, while RPT penalizes the sum of the eigenvalues~\cite{abbas2016understanding}.
Therefore, so long that the \textit{true} output distribution is isotropically smooth around the input data point, VAT tends to be more effective in improving the generalization performance.

In Section~\ref{subsec:VATvsRPT}, we investigate the variance of the gradients in more detail and compare RPT and VAT from this perspective. 

\section{\label{sec:experiments}Experiments}
We conducted a set of numerical experiments to assess the following aspects of VAT:  
\begin{tight_itemize}
\item the sheer efficacy of VAT in comparison to RPT and to a collection of recent competitive algorithms for supervised and semi-supervised learning,
\item the effect of hyperparameters (the perturbation size $\epsilon$, the regularization coefficient $\alpha$, and the number of the power iterations $K$) on the performance of VAT, 
\item VAT's effect on the robustness of the trained NNs against virtual adversarial perturbations, and
\item the mechanism behind the advantage of using virtual adversarial perturbation as opposed to random perturbation.
\end{tight_itemize} 
We would like to remind the readers that, by the term VAT here, we mean the algorithm that uses the approximation step we introduced in Section~\ref{subsec:fastapprox} and Algorithm~\ref{algo:vat_loss}. In the following, we describe the experimental settings and outcomes of the experiments. 
For the performance evaluation of our method, we used standard benchmarks like MNIST, CIFAR-10 and SVHN.  
For the methods to compare, we used the methods that were state-of-the-art at the time of this research.
For more details on the dataset and model architectures, please see the appendix sections.
For all our experiments, we used fully connected NNs or convolutional neural networks~(CNNs) as the architectures of the classifiers, and 
used Theano~\cite{2016arXiv160502688short} and TensorFlow~\cite{abadi2016tensorflow} to train the models\footnote{TensorFlow implementation for the experiments is available at \url{https://github.com/takerum/vat_tf}. 
Chainer\cite{tokui2015chainer} implementation is also available at \url{https://github.com/takerum/vat_chainer}.}.
We use $p(y|x,\theta)$ to denote the label distribution of the classifier, where $\theta$ represents the vector of the parameters of the NN.
For the activation function, we used ReLU~\cite{jarrett2009best, nair2010rectified, glorot2011deep}.  
We also used Batch Normalization~\cite{ioffe2015batch}. 
For the divergence $D$ in Eq.~\eqref{eq:adv_loss}, we chose the KL divergence 
\begin{eqnarray}
	D\left[p(y|x), p(y|x+r)\right] \defeq \sum_{y\in Q} p(y|x) \log \frac{p(y|x)}{p(y|x+r)},
\end{eqnarray}
where $Q$ is the domain of $y$. For classification problems, $Q$ is the set of all possible labels. 
We set $\xi$ in Eq.\eqref{eq:dif_method} to be $1e\text{-}6$ in all of our experiments. 
For each set of experiments, we repeated the same procedure multiple times with different random seeds for the weight initialization and for the selection of labeled samples (for semi-supervised learning). 
The values reported on the tables are the means and the standard deviations of the results. 
Please see the appendix sections for more details.

\subsection{\label{subsec:efficacy}Testing the Efficacy of VAT on Benchmark Tasks}
\subsubsection{Supervised Learning on MNIST and CIFAR-10}
We first applied our algorithm to supervised learning on MNIST dataset. We used NNs with four hidden layers, whose numbers of units were $(1200, 600, 300, 150)$. 
We provide more details of the experiments in Appendix~\ref{apd:mnist_exp_set}. 

For each regularization method, we used the set of hyperparameters that achieved the best performance on the validation dataset of size $10,000$, which was selected from the pool of training samples of size $60,000$. 
Test datasets were produced so that they have no intersection with the training dataset.
We fed the test dataset into the trained NNs and recorded their test errors. 

\begin{figure}[ht]
	\centering
	\includegraphics[width=0.45\textwidth]{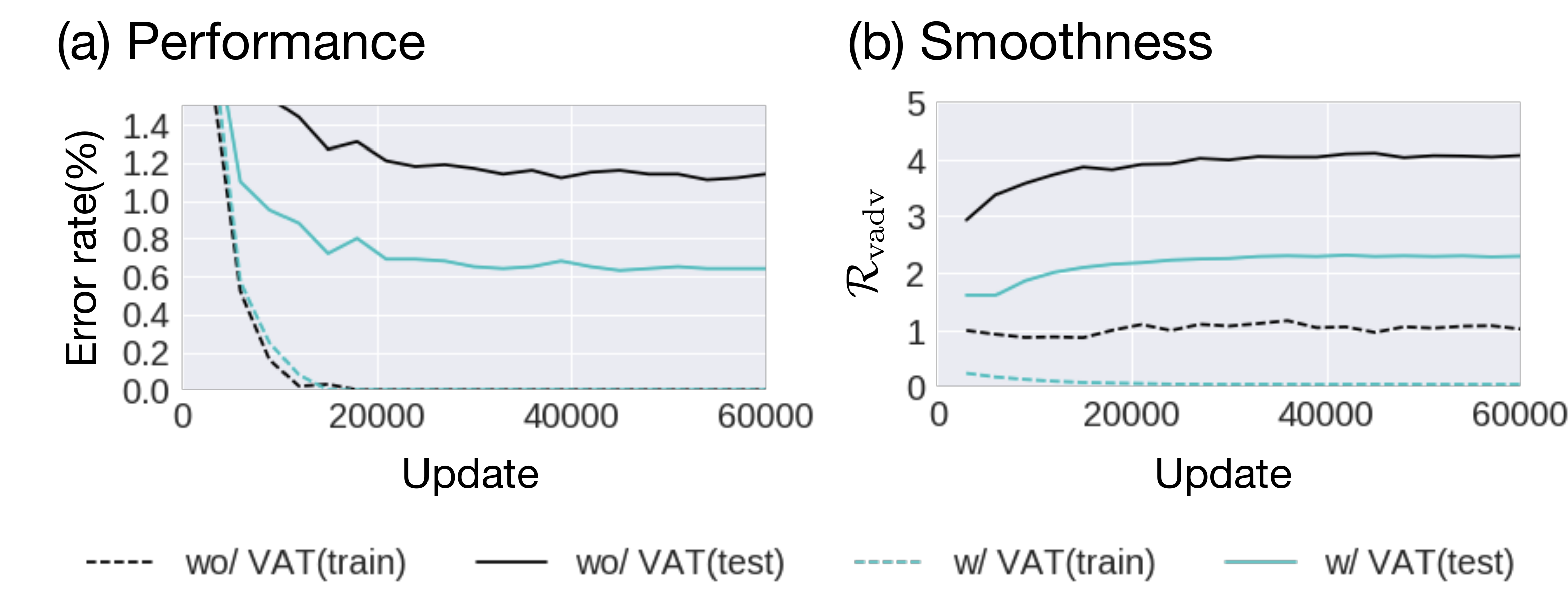}
	\caption{\label{fig:mnist_lc}Transition of the (a) classification error and (b) $\mathcal{R}_{\rm vadv}$ for supervised learning on MNIST. 
    We set $\epsilon=2.0$ for the evaluation of $\mathcal{R}_{\rm vadv}$ for both the baseline and VAT.
    This is the value of $\epsilon$ with which the VAT-trained model achieved the best performance on the validation dataset.}
\end{figure}

Fig. \ref{fig:mnist_lc} shows the transition of $\mathcal{R}_{\rm vadv}$ and the learning curves for the baseline NN trained \textit{without} VAT (denoted by `wo/ VAT') and the NN trained \textit{with} VAT (denoted by `w/ VAT'). 
As the training progressed, $\mathcal{R}_{\rm vadv}$ of the NN trained with VAT exceeded that of the baseline; that is, 
the model trained with VAT grew smoother than the baseline in terms of LDS.  

Table \ref{tab:supMNIST} summarizes the performance of our regularization method (VAT) and the other regularization methods for supervised learning on MNIST. VAT performed better than all the contemporary methods except ladder networks, which is a highly advanced method based on special network structure. 

We also tested VAT with $K >1$ in order to study the dependence of the number of the power iterations $K$ on the performance of VAT. 
As we will discuss in Section~\ref{subsec:piter}, however, we were not able to achieve substantial improvement by increasing the value of $K$. 

We also applied our algorithm to supervised learning on CIFAR-10 ~\cite{krizhevsky2009learning}. 
For the baseline for this set of experiments, we used a `Conv-Large' with dropout~(Table~\ref{tab:cnn_models}, Appendix~\ref{apd:semisup_details}).
Table \ref{tab:supCIFAR-10} summarizes the test performance of supervised learning methods implemented with CNN on CIFAR10. We also compared the performance of VAT to advanced architectures like ResNet~\cite{he2016identity} and DenseNet~\cite{huang2016densely} in order to confirm that the baseline model of our algorithm is ``mediocre enough" so that we can rightly attribute the effectiveness of our algorithm to its way of the  regularization itself, not to the network structure we used in the experiments.
For CIFAR-10, our VAT achieved satisfactory performance relative to the standards.  

\begin{table}[ht]
  \centering
		\caption{\label{tab:supMNIST}Test performance of supervised learning methods on MNIST with 60,000 labeled examples in the permutation invariant setting.  
        The top part cites the results provided by the original paper. 
        The bottom part shows the performance achieved by our implementation.  }
		\begin{tabular}{lr}
			\toprule
			Method & Test error rate(\%)  \\
			\midrule
            SVM (Gaussian kernel) & 1.40 \\
            Dropout \cite{srivastava2014dropout} & 1.05 \\
            Adversarial, $L_{\infty}$ norm constraint \cite{goodfellow2014explaining} & 0.78\\
            Ladder networks \cite{rasmus2015semi}& \textbf{0.57} ($\pm$0.02)\\
            \midrule
            Baseline (MLE) & 1.11 ($\pm$0.06)\\
            RPT & 0.84 ($\pm$0.03) \\
            Adversarial, $L_{\infty}$ norm constraint & 0.79 ($\pm$0.03)\\ 
            Adversarial, $L_2$ norm constraint & 0.71 ($\pm$0.03)\\ 
            VAT & 0.64 ($\pm$0.05) \\
			\bottomrule
		\end{tabular}
\end{table}

\begin{table}[ht]
  \centering
		\caption{\label{tab:supCIFAR-10}Test performance of supervised learning methods implemented with CNN on CIFAR-10 with 50,000 labeled examples.
        The top part cites the results provided by the original paper. 
        The bottom part shows the performance achieved by our implementation.}
		\begin{tabular}{lr}
			\toprule
			Method & Test error rate(\%)  \\
			\midrule
            Network in Network~\cite{lin2013network}  & 8.81 \\
            All-CNN~\cite{springenberg2014striving}  & 7.25 \\
            Deeply Supervised Net~\cite{lee2015deeply}  & 7.97\\
            Highway Network~\cite{srivastava2015highway} & 7.72\\
            ResNet (1001 layers)~\cite{he2016identity}&　4.62~($\pm$0.20)\\
            DenseNet (190 layers)~\cite{huang2016densely} & \textbf{3.46} \\
            \midrule
            Baseline (only with dropout) & 6.67 ($\pm$0.07) \\ 
            RPT & 6.30 ($\pm$0.04) \\ 
            VAT & 5.81 ($\pm$0.02)  \\ 
			\bottomrule
		\end{tabular}
\end{table}

\subsubsection{\label{subsec:semisup}Semi-Supervised Learning on MNIST, SVHN, and CIFAR-10}
Recall that our definition of ${\rm LDS}(x, \theta)$ at any point $x$ does not require the supervisory signal for $x$.
In particular, this means that we can apply VAT to semi-supervised learning tasks. 
We emphasize that this is a property not shared by adversarial training.
We applied VAT to semi-supervised image classification tasks on three datasets: MNIST, SVHN \cite{netzer2011reading}, and CIFAR-10 \cite{krizhevsky2009learning}. 
We provide the details of the experimental settings in Appendix \ref{apd:semisup_details}.

For MNIST, we used the same architecture of NN as in the previous section. We also used batch normalization in our implementation.
We used a mini-batch of size 64 for the calculation of the negative log-likelihood term, and a mini-batch of size 256 for the calculation of $\mathcal{R}_{\rm vadv}$ in Eq. \eqref{eq:full}. As mentioned in Section \ref{subsec:vat}, we used both labeled and unlabeled sets for the calculation of  $\mathcal{R}_{\rm vadv}$. 
Table \ref{tab:semisupMNIST} summarizes the results for the permutation invariant MNIST task. All the methods listed in the table belong to the family of semi-supervised learning methods. For the MNIST dataset, VAT outperformed all the contemporary methods other than the methods based on generative models, such as \cite{rasmus2015semi,maaloe2016auxiliary,salimans2016improved}, which are state-of-the-art methods.
\begin{table}[ht]
  \centering
		\caption{\label{tab:semisupMNIST}Test performance of semi-supervised learning methods on MNIST with the permutation invariant setting.
        The value $N_l$ stands for the number of labeled examples in the training set.   
        The top part cites the results provided by the original paper. 
        The bottom part shows the performance achieved by our implementation. 
        (PEA = Pseudo Ensembles Agreement, DGM = Deep Generative Models, FM=feature matching) }
		\begin{tabular}{lrr}
			\toprule
			\multirow{2}{*}{Method} & \multicolumn{2}{c}{Test error rate(\%)}  \\
		     & $N_l=100$ &  $N_l=1000$ \\
			\midrule
            TSVM \cite{collobert2006large} & 16.81 & 5.38\\
            PEA \cite{bachman2014learning} & 5.21 & 2.87\\ 
            DGM (M1+M2) \cite{kingma2014semi} & 3.33 ($\pm$0.14) & 2.40 ($\pm$0.02)\\

            CatGAN \cite{springenberg2015unsupervised} & 1.91 ($\pm$0.1)& 1.73 ($\pm$0.18) \\
            Skip DGM \cite{maaloe2016auxiliary} & 1.32 ($\pm$0.07) &  \\
            Ladder networks~\cite{rasmus2015semi} & 1.06 ($\pm$0.37) & \textbf{0.84} ($\pm$0.08)\\
			Auxiliary DGM~\cite{maaloe2016auxiliary}& 0.96 ($\pm$0.02)& \\
            GAN with FM \cite{salimans2016improved}& \textbf{0.93} ($\pm$0.07) & \\
            \midrule
            RPT & 6.81 ($\pm$1.30) & 1.58 ($\pm$0.54)\\ 
            VAT & 1.36 ($\pm$0.03) & 1.27 ($\pm$0.11) \\ 
			\bottomrule
		\end{tabular}
\end{table}

For the experiments on SVHN and CIFAR-10, we used two types of CNNs (Conv-Small and Conv-Large) used in recent state-of-the-art semi-supervised learning methods (\cite{rasmus2015semi,springenberg2015unsupervised,salimans2016improved,laine2016temporal}). 
`Conv-Small CNN' has practically the same structure as the CNN used in \cite{salimans2016improved}, and `Conv-Large CNN' has practically the same structure as the CNN used in \cite{laine2016temporal}.
We provide more details of the architectures in Table~\ref{tab:cnn_models} of Appendix~\ref{apd:semisup_details}.

Also, in SVHN and CIFAR-10 experiments, we adopted conditional entropy of $p(y|x,\theta)$ as an additional cost: 
\begin{align}
	\mathcal{R}_{\rm cent} &= \mathcal{H}(Y|X) \nonumber \\
    &= -\frac{1}{N_l + N_{ul}}\sum_{x\in \mathcal{D}_l, \mathcal{D}_{ul}}\sum_{y} p(y|x, \theta) \log p(y|x, \theta).
\end{align}
This cost was introduced by \cite{grandvalet2004semi}, and similar idea has been used in \cite{sajjadi2016regularization}.
The conditional entropy minimization has an effect of exaggerating the prediction of the model $p(y|x,\theta)$ on each data point.
For semi-supervised image classification tasks, this additional cost is especially  helpful.  
In what follows, `VAT+EntMin' indicates the training with $\mathcal{R}_{\rm vadv} + \mathcal{R}_{\rm cent}$.

Table~\ref{tab:semisup_wodataaug} summarizes the results of semi-supervised learning tasks on SVHN and CIFAR-10.
Our method achieved the test error rate of $14.82$(\%) with VAT, which outperformed the state-of-the-art methods for semi-supervised learning on CIFAR-10. 
`VAT+EntMin' outperformed the state-of-the-art methods for semi-supervised learning on both SVHN and CIFAR-10.

\begin{table}[ht]
  \centering
		\caption{\label{tab:semisup_wodataaug} Test performance of semi-supervised learning methods on SVHN and CIFAR-10 \textit{without} image data augmentation. 
        The value $N_l$ stands for the number of labeled examples in the training set. 
        The top part cites the results provided by the original paper. 
        The middle and bottom parts show the performance achieved by our implementation. 
        The asterisk(*) stands for the results on the permutation invariant setting. (DGM=Deep Generative Models, FM=feature matching)
}
		\begin{tabular}{lrr}
			\toprule
			\multirow{3}{*}{Method} & \multicolumn{2}{c}{Test error rate(\%) }  \\
		     & SVHN & CIFAR-10 \\
             & $N_l=1000$ & $N_l=4000$ \\
			\midrule
            SWWAE \cite{zhao2015stacked}& 23.56 & \\
            *Skip DGM~\cite{maaloe2016auxiliary} & 16.61 ($\pm$0.24) &  \\
            *Auxiliary DGM~\cite{maaloe2016auxiliary} & 22.86 &  \\
			Ladder networks, $\Gamma$ model \cite{rasmus2015semi} &  & 20.40 ($\pm$0.47) \\
            CatGAN \cite{springenberg2015unsupervised} & & 19.58 ($\pm$0.58) \\
            GAN with FM \cite{salimans2016improved}& 8.11 ($\pm$1.3) &18.63 ($\pm$2.32) \\
			$\prod$ model \cite{laine2016temporal} & 5.43 ($\pm$0.25) & 16.55 ($\pm$0.29)\\

            \midrule
            (on Conv-Small used in \cite{salimans2016improved})\\
            RPT & 8.41 ($\pm$0.24) & 18.56 ($\pm$0.29)  \\
            VAT & \textbf{6.83} ($\pm$0.24)  & \textbf{14.87} ($\pm$0.13) \\
            \midrule
            (on Conv-Large used in \cite{laine2016temporal})\\
            VAT & 5.77 ($\pm$0.32)  &  14.18 ($\pm$0.38) \\
            VAT+EntMin & \textbf{4.28} ($\pm$0.10)  & \textbf{13.15} ($\pm$0.21) \\
			\bottomrule
		\end{tabular}
\end{table}

Table~\ref{tab:semisup_dataaug} shows the performance of VAT and contemporary semi-supervised learning methods implemented with moderate image data augmentation (translation and horizontal flip).
All methods other than VAT were implemented with the strong standing assumption on the unlabeled training samples that the label of the image does not change by the deformation. 
On CIFAR-10, VAT still outperformed the listed methods with this handicap. `VAT+EntMin' with moderate data augmentation also outperformed the current state-of-the-art methods for semi-supervised learning on both SVHN and CIFAR-10. 
This result tells us that the effect of VAT does not overlap so much with that of the data augmentation methods, so that they can be used in combination to boost the performance.

\begin{table}[ht]
  \centering
		\caption{\label{tab:semisup_dataaug} Test performance of semi-supervised learning methods on SVHN and CIFAR-10 \textit{with} image data augmentation. 
        The value $N_l$ stands for the number of labeled examples in the training set. The performance of all methods other than Sajjadi et al. \cite{sajjadi2016regularization} are based on experiments with the moderate data augmentation of translation and flipping (see Appendix~\ref{apd:semisup_details} for more detail).
        Sajjadi et al. \cite{sajjadi2016regularization} used extensive image augmentation, which included rotations, stretching, and shearing operations. 
        The top part cites the results provided by the original paper. 
        The bottom part shows the performance achieved by our implementation.
        }
		\begin{tabular}{lrr}
			\toprule
			\multirow{3}{*}{Method} & \multicolumn{2}{c}{Test error rate(\%)}  \\
		     & SVHN & CIFAR-10 \\
             & $N_l=1000$ & $N_l=4000$ \\
\midrule
			$\prod$ model \cite{laine2016temporal}. & 4.82 ($\pm$0.17) & 12.36 ($\pm$0.31)\\
            Temporal ensembling \cite{laine2016temporal} & 4.42 ($\pm$0.16)& 12.16 ($\pm$0.24) \\
            Sajjadi et al. \cite{sajjadi2016regularization} & & 11.29 ($\pm$0.24)\\
            \midrule
            (On Conv-Large used in \cite{laine2016temporal})\\
            VAT & 5.42 ($\pm$0.22) & 11.36 ($\pm$0.34) \\
            VAT+EntMin & \textbf{3.86} ($\pm$0.11) & \textbf{10.55} ($\pm$0.05) \\ 
			\bottomrule
		\end{tabular}
\end{table}

\subsection{\label{subsec:hp}
Effects of Perturbation Size $\epsilon$ and Regularization Coefficient $\alpha$}
\noindent Generally, there is a high computational cost for optimizing the hyperparameters of large NNs on a large dataset.  
One advantage of VAT is that the algorithm involves only two hyperparameters: $\alpha$ and $\epsilon$.
Even better, our experiments suggest that, in an appropriate setting, VAT can be made effective with the optimization of $\epsilon$ alone.
In all our experiments in Section~\ref{subsec:efficacy}, VAT achieved competitive results while fixing $\alpha =1$. 

This result is not too counter-intuitive. For small $\epsilon$, the hyperparameter $\alpha$ plays a similar role as $\epsilon$. To see this, consider the Taylor expansion of Eq.\eqref{eq:vadv_loss} for small $\epsilon$, given by
\begin{align}
\max_r\{D(r, x, \theta); \|r\|_2 \leq \epsilon\} &\approx \max_r\{\frac{1}{2}r^{T}H(x,\theta)r; \|r\|_2 \leq \epsilon\}\nonumber\\
				&=\frac{1}{2} \epsilon^2 \lambda_1(x, \theta),
\end{align}
where $H(x,\theta)$ and $\lambda_1(x, \theta)$ are respectively the Hessian matrix of $D$ in Eq.\eqref{eq:second_taylor} and its dominant eigenvalue.
Substituting this into the objective function (Eq.\eqref{eq:full}), we obtain
\begin{align}
&\ell(\theta,\mathcal{D}_l) +  \alpha \mathcal{R}_{\rm vadv}(\theta, \mathcal{D}_l, \mathcal{D}_{ul}, ) \nonumber \\
=&\ell(\theta,\mathcal{D}_l) + \alpha\frac{1}{N_l+N_{ul}}\sum_{x_* \in \mathcal{D}_l, \mathcal{D}_{ul}}\max_r\{D(r, x_*, \theta),
\|r\|_2 \leq \epsilon\} \nonumber \\ 
\approx &\ell(\theta,\mathcal{D}_l) + \frac{1}{2}\alpha\epsilon^2 \frac{1}{N_l+N_{ul}}\sum_{x_* \in \mathcal{D}_l, \mathcal{D}_{ul}}\lambda_1(x_*, \theta).  \label{eq:approx_ep} 
\end{align}
Thus, at least for small $\epsilon$, the strength of the regularization in VAT is proportional to the product of the two hyperparameters, $\alpha$ and $\epsilon^2$; 
that is, in the region of small $\epsilon$, the hyperparameter search for only either $\epsilon$ or $\alpha$ suffices. However, when we consider a relatively large value of $\epsilon$, the hyperparameters $\alpha$ and $\epsilon$ cannot be brought together.
In such a case, we shall strive to search for the best pair of hyperparameters that attains optimal performance.

In our experiments on MNIST, tuning of $\epsilon$ alone sufficed for achieving satisfactory performance.
We therefore recommend on empirical basis that the user prioritizes the parameter search for $\epsilon$ over the search for $\alpha$. 

\begin{figure}[ht]
	\centering
	\begin{subfigure}{0.24\textwidth}
	\includegraphics[width=\textwidth]{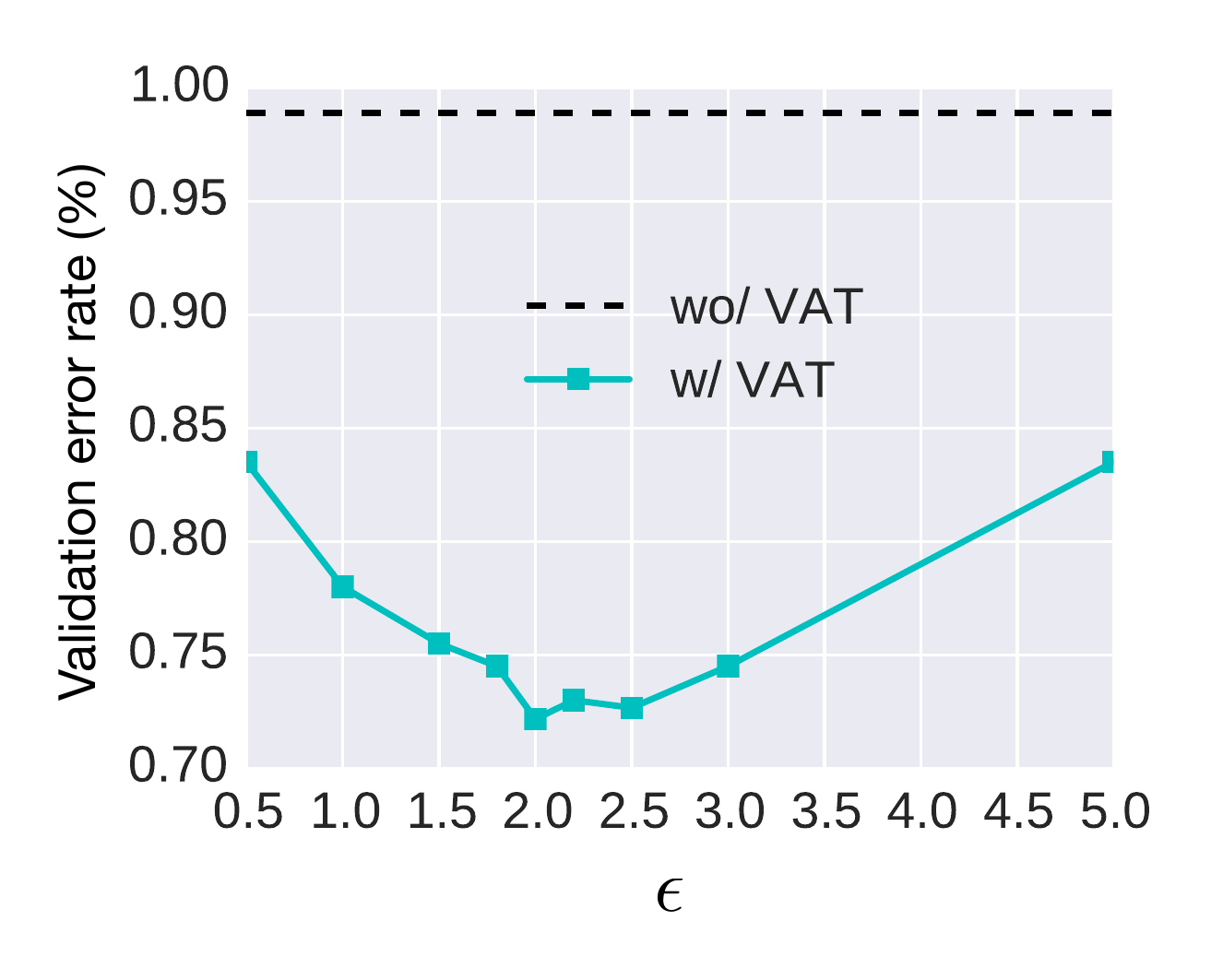}
	\caption{\label{fig:mnist_epsilon}Effect of $\epsilon$ ($\alpha=1$).}
    \end{subfigure}
    \begin{subfigure}{0.24\textwidth}
	\includegraphics[width=\textwidth]{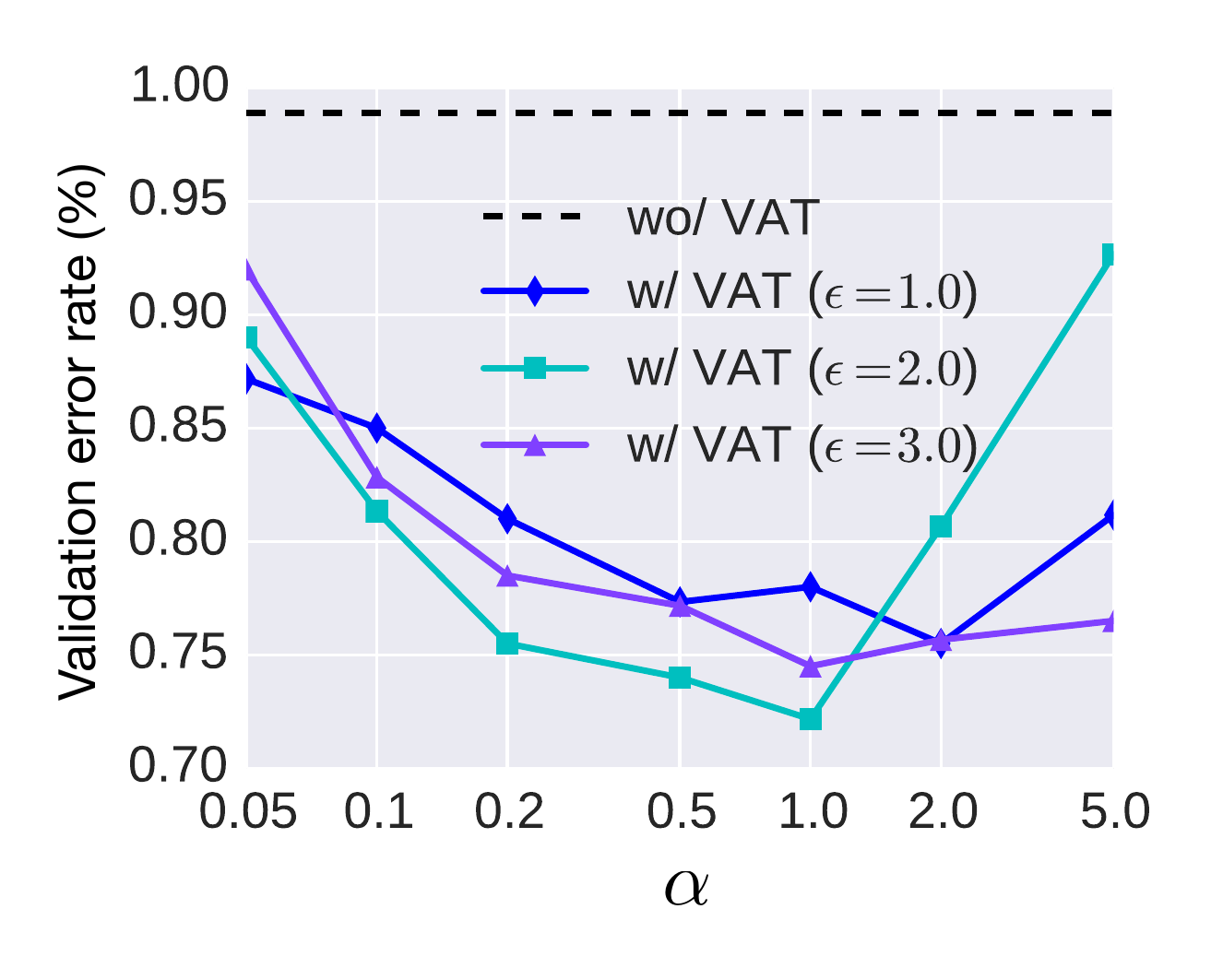}
	\caption{\label{fig:mnist_lambda}Effect of $\alpha$.}
    \end{subfigure}
   	\caption{\label{fig:mnist_ep_lambda} Effect of $\epsilon$ and $\alpha$ on the validation performance for supervised task on MNIST.}
\end{figure}

Fig. \ref{fig:mnist_ep_lambda} shows the effects of $\epsilon$ and $\alpha$ on the validation performance of supervised learning on MNIST.
In Fig. \ref{fig:mnist_epsilon}, we show the effect of $\epsilon$ with fixed $\alpha=1$, and in Fig. \ref{fig:mnist_lambda} we show the effects of $\alpha$ with different fixed values of $\epsilon$ in the range $\{1.0, 2.0, 3.0\}$.
During the parameter search for supervised learning on MNIST, the algorithm performed optimally when $\alpha =1$.
Based on this result, we fixed $\alpha =1$ while searching for optimal $\epsilon$ in all benchmark experiments, including both supervised learning on MNIST and CIFAR-10, as well as  unsupervised learning on  MNIST, CIFAR-10, and SVHN.
As we showed in Section~\ref{subsec:efficacy}, this simple tuning of $\epsilon$  was sufficient for good performance of VAT, and it even achieved state-of-the-art performance for several tasks. 

\subsection{\label{subsec:piter} Effect of the Number of the Power Iterations $K$}

\begin{figure}[ht]
	\centering
	\begin{subfigure}{0.24\textwidth}
	\includegraphics[width=\textwidth]{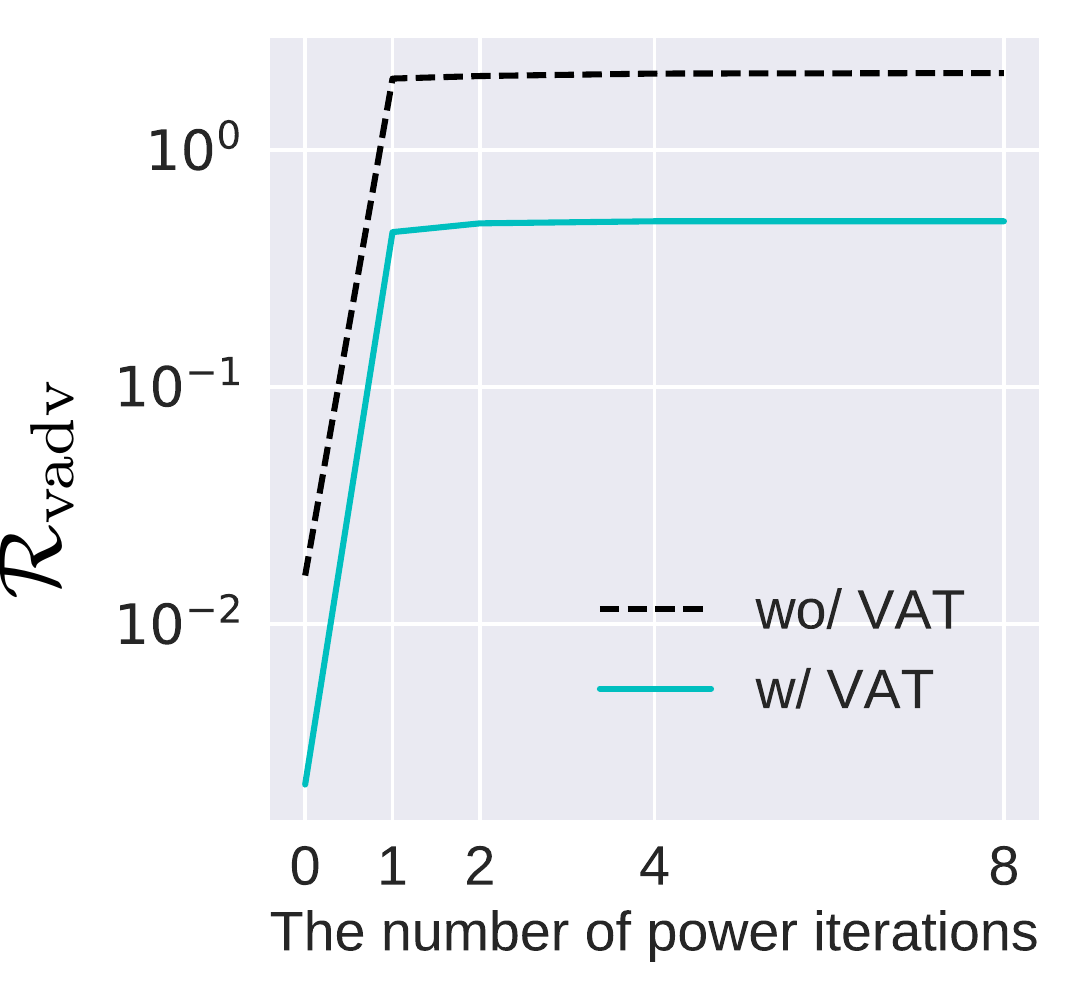}
	\caption{\label{fig:mnist_powiter}MNIST}
    \end{subfigure}
    \begin{subfigure}{0.24\textwidth}
	\includegraphics[width=\textwidth]{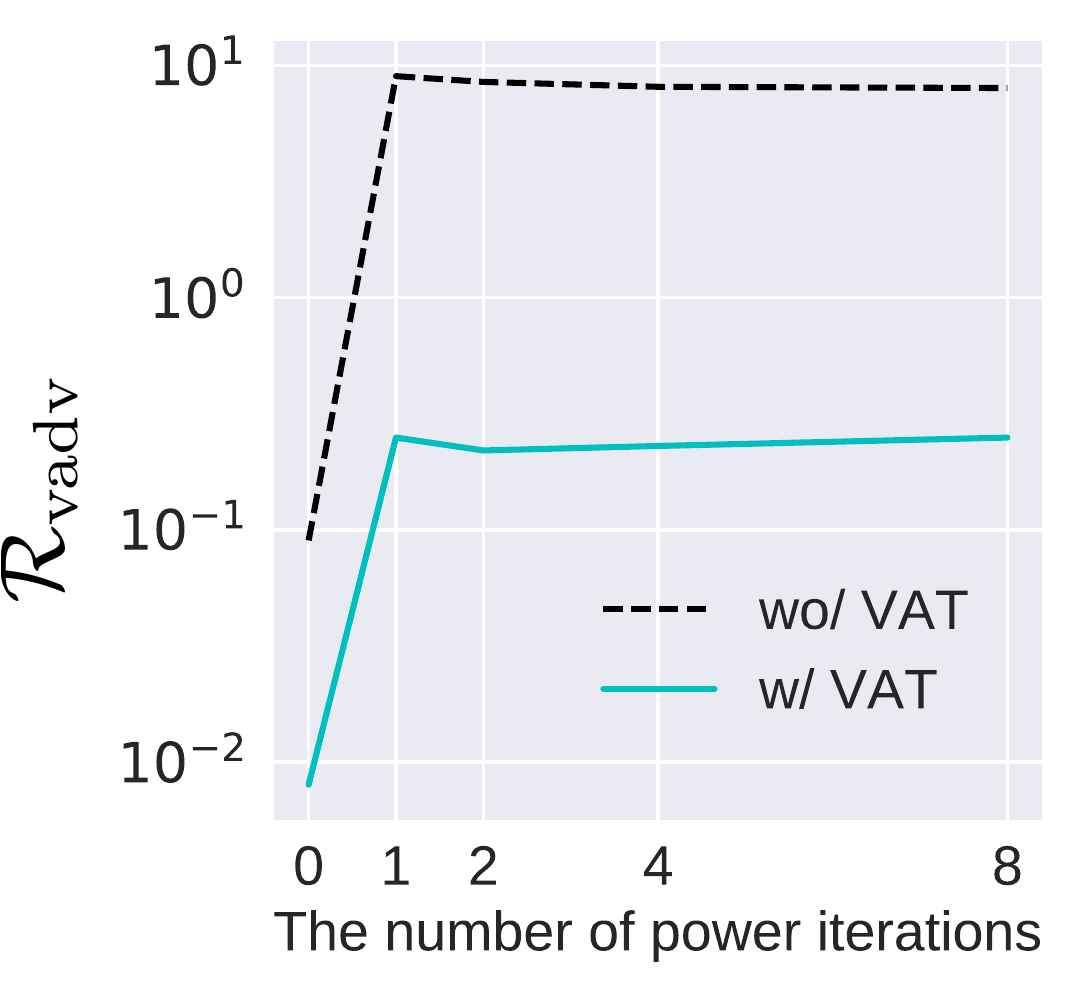}
	\caption{\label{fig:cifar10}CIFAR-10}
    \end{subfigure}
    \caption{\label{fig:powiter}Effect of the number of the power iterations on $\mathcal{R}_{\rm vadv}$ for (a) supervised task on MNIST and (b) semi-supervised task on CIFAR-10.}
\end{figure}

Fig. \ref{fig:powiter} shows the $\mathcal{R}_{\rm vadv}$ values of the models trained with different $K$~(the number of the power iterations) for supervised learning on MNIST and semi-supervised learning on CIFAR-10. We can observe a significant increase in the value of $\mathcal{R}_{\rm vadv}$ over the transition from $K=0$ (random perturbations) to $K=1$ (virtual adversarial perturbations). We 
can also observe that the value saturates at $K =1$. 
The fact that one power iteration sufficed for good performance tells that the ratio $\lambda_1/\lambda_2$ for our experimental settings was very large.

In fact, we could not achieve notable improvement in performance by increasing the value of $K$. 
Table~\ref{tab:semisup_cifar10_poweriter} shows the test accuracies for the semi-supervised learning task on CIFAR10 with different values of $K$.
We however shall note that there is no guarantee that the spectrum of Hessian is always skew. Depending on the dataset and the model,  $K=1$ might not be sufficient. 

\begin{table}[ht]
  \centering
		\caption{\label{tab:semisup_cifar10_poweriter} 
        The test accuracies of VAT for the semi-supervised learning task on CIFAR10 with different values of $K$ (the number of the power iterations). }
		\begin{tabular}{lrr}
			\toprule
			\multirow{2}{*}{} & \multicolumn{1}{c}{Test error rate(\%)}  \\
		     & CIFAR-10 \\
             & $N_l=4000$ \\
\midrule
            (On Conv-Large)\\
			VAT, $K=1$ & 14.18 ($\pm$0.38)\\
            VAT, $K=2$ & 14.19 ($\pm$0.16)\\
            VAT, $K=4$ & 14.25 ($\pm$0.18)\\
			\bottomrule
		\end{tabular}
\end{table}


\subsection{Visualization of Virtual Adversarial Examples}
\subsubsection{Virtual Adversarial Examples Produced by the Model Trained with Different Choices of $\epsilon$}
\begin{figure*}
	\centering
    \hspace*{\fill}%
    \begin{subfigure}{0.35\textwidth}
 	\includegraphics[width=\textwidth]{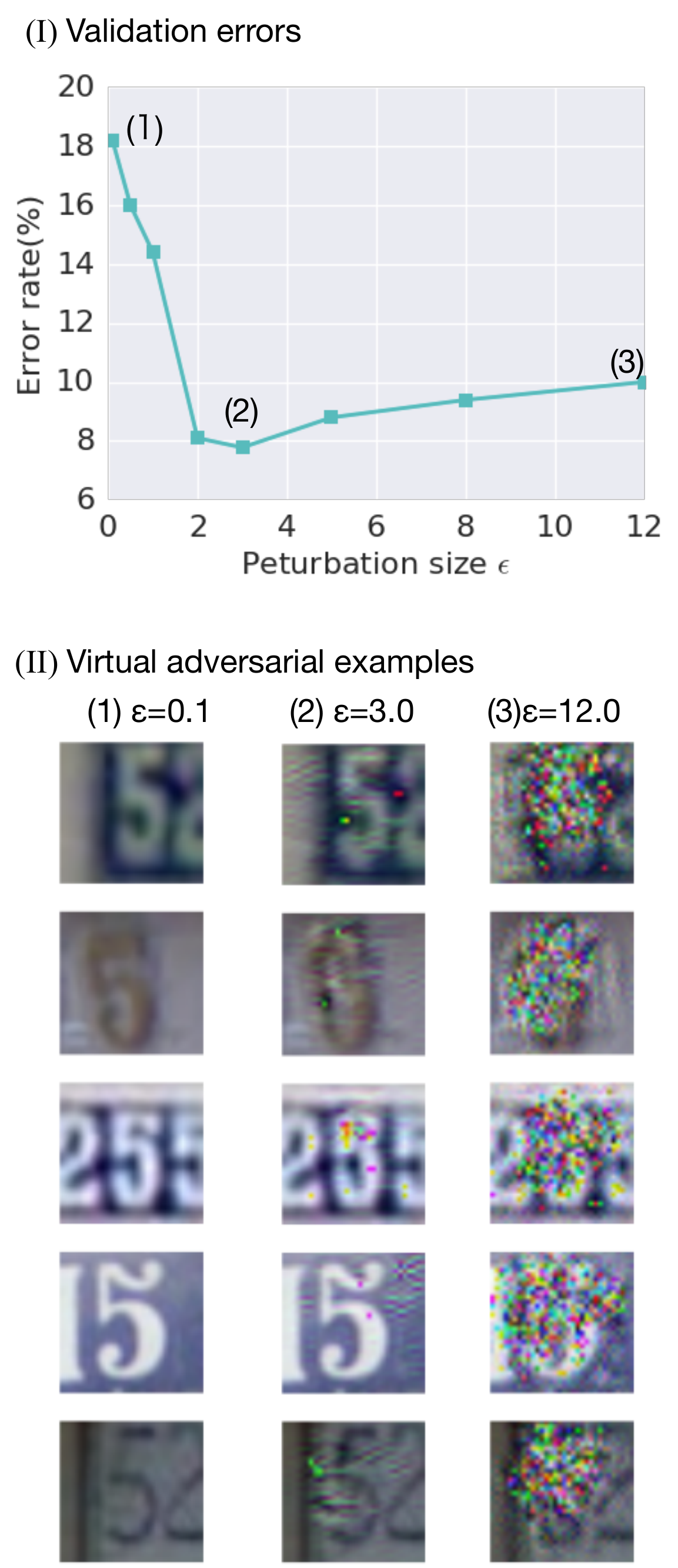}
	\caption{\label{fig:svhn_epsilon_vae}SVHN}
    \end{subfigure}\hfill%
    \begin{subfigure}{0.35\textwidth}
	\includegraphics[width=\textwidth]{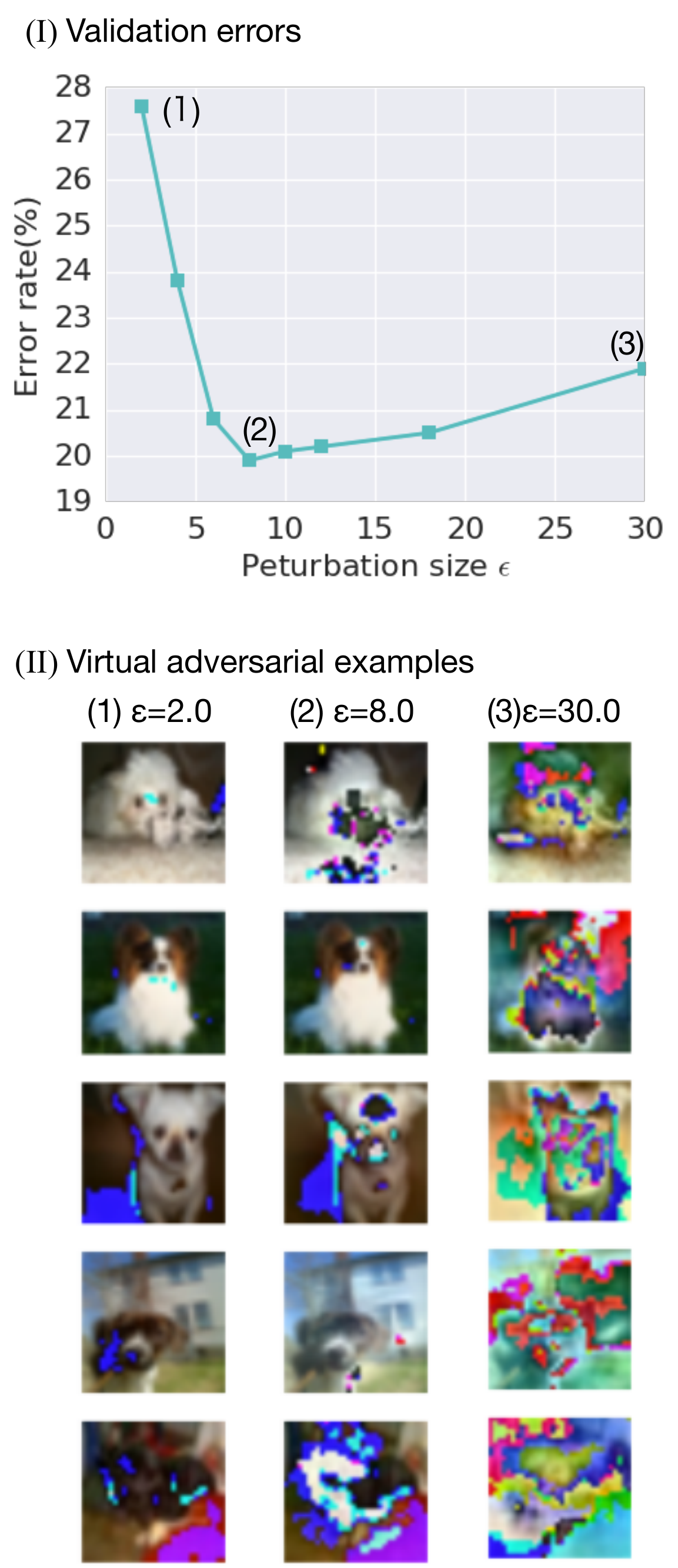}
	\caption{\label{fig:cifar10_epsilon_vae}CIFAR-10}
    \end{subfigure}
    \hspace*{\fill}%
    \caption{\label{fig:epsilon_vae} \textbf{Performance of VAT with different values of $\epsilon$.}   The effect of $\epsilon$ on the performance of semi-supervised learning　(I), together with the set of typical virtual adversarial examples generated by the model trained with VAT with the corresponding value of $\epsilon$ (II). }
\end{figure*}

One might be interested in the actual visual appearance of the virtual adversarial examples with an appropriate size of $\epsilon$ that improves the regularization performance. 
In Fig. \ref{fig:epsilon_vae}, we aligned (I) the transition of the performance of VAT on SVHN and CIFAR-10 with respect to $\epsilon$ along (II) the actual virtual adversarial examples that the models trained with corresponding $\epsilon$ generated at the end of its training. 
For small $\epsilon$ (designated as (1) in the figure), it is difficult for human eyes to distinguish the virtual adversarial examples from the clean images. 
The size of $\epsilon$ that achieved the best validation performance is designated by (2).  
Especially for CIFAR-10, the virtual adversarial examples with $\epsilon$ of size (2) are on the verge of total corruption.   
For a larger value of $\epsilon$ (designated by (3) in the figure), we can clearly observe the effect of over-regularization. 
In fact, the virtual adversarial examples generated by the models trained with this range of $\epsilon$ are very far from the clean image, and we observe that the algorithm implemented with this large an $\epsilon$ did the unnecessary work of smoothing the output distribution over the set of images that are ``unnatural." 

\subsubsection{Robustness against Virtual Adversarial Examples after Training}

We studied the nature of the robustness that can be attained by VAT.
We trained CNNs on CIFAR-10 with and without VAT and prepared a set of pairs of virtual adversarial examples generated from the same picture, each consisting of (1) the virtual adversarial example generated by the model trained with VAT~(w/ VAT) and (2) the virtual adversarial example generated by the model trained without VAT~(wo/ VAT).
We studied the rates of misidentification by the classifiers~(w/VAT and wo/VAT) on these pairs of adversarial examples.

\begin{figure*}
	\centering   
    	\includegraphics[width=0.8\textwidth]{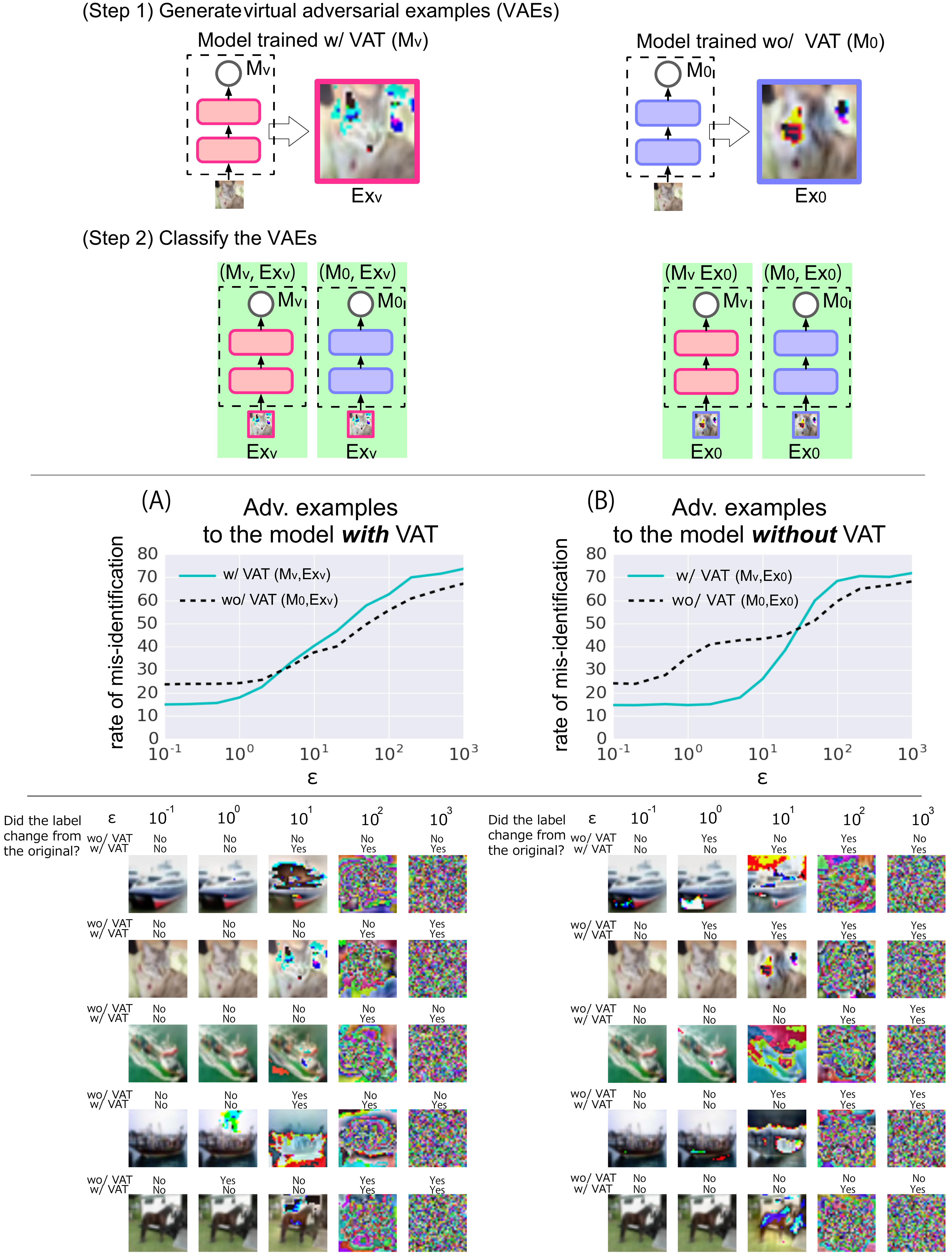}
		\caption{\label{fig:misid}\textbf{Robustness of the VAT-trained model against perturbed images.}  
        The upper panel shows the procedure for the evaluation of robustness.  
        In step 1, we prepared two classifiers -- one trained with VAT (${\rm M}_v$)  and another trained without VAT (${\rm M}_0$) -- and generated a virtual adversarial example from each classifier (${\rm Ex}_v$ and ${\rm Ex}_0$). In step 2, we classified ${\rm Ex}_v$ and ${\rm Ex}_0$ by these two models, thereby yielding the total of four classification results. 
    	The middle panel (graph A and B) plots the rate of misidentification  in these four classification tasks against the size of the perturbation ($\epsilon$) used to generate the virtual adversarial examples~(VAEs) in step 1. 
        The left half of the bottom panel aligned with the graph (A) shows a set of ${\rm Ex}_v$ generated with different values of $\epsilon$, together with the classification results of ${\rm M}_v$ and ${\rm M}_0$ on the images. 
        All ${\rm Ex}_v$ listed here are images generated from a set of clean examples that were \textit{correctly} identified by ${\rm M}_v$ and ${\rm M}_0$. 
        The right half of the bottom panel aligned with graph (B) shows the set of ${\rm Ex}_0$ generated from the same clean images as ${\rm Ex}_v$ with different values of $\epsilon$. 
        The label `Yes' indicates that the model changed the label assignment when the perturbation was applied to the image~(e.g. the model was deceived by the perturbation)
        The label 'No' indicates that the model maintained the label assignment on the perturbed image.
		Note that the model ${\rm M}_v$ dominates the model ${\rm M}_0$ in terms of classification performance on the images that appear almost indistinguishable from the clean image. }
\end{figure*}

Fig.~\ref{fig:misid} shows the rates at which the two models~(w/VAT and wo/VAT) misidentified the images corrupted by virtual adversarial perturbations of different magnitudes. 
The figure (A) in the middle panel shows the rates of misidentification made on the virtual adversarial examples generated by the model trained with VAT. The figure (B) shows the rates on the virtual adversarial examples generated by the model trained without VAT. 
The example pictures shown beneath the figures (A) and (B) are the adversarial examples generated from the set of images that were correctly identified by both the model trained with VAT and the model trained without VAT when fed without perturbation.
As expected, the error rates increased monotonically with the intensity of corruption for both models. 
Overall, we recognize that the adversarial examples generated by both models are almost identical to the original image for human eyes when $\epsilon \sim 10^{-1}$.
The adversarial examples around $\epsilon \sim 10^0$ are almost identifiable, but are so corrupted that any further corruption would make the image unidentifiable by human eyes.
The virtual adversarial examples with this range of $\epsilon$ are therefore the examples on which we wish the classifier to make no mistakes. 
We can clearly observe that the rate of misidentification made by the VAT-trained model for this range of $\epsilon$ is much lower than that of the model trained without VAT. 
Also note in the bottom panel that the model trained with VAT correctly identifies both the adversarial examples generated by itself and the adversarial examples generated by the model trained without VAT for this range of $\epsilon$.  
Simultaneously, note that the model trained with VAT alters its decision on the images when the perturbation is too large.
In contrast, the model trained without VAT is assigning the original labels to the over-perturbed images at much higher rate than the VAT-trained model, which is completely unnecessary, and is possibly even harmful in practice.
Thus, we observe that the VAT-trained model behaves much more `naturally' than the model trained without VAT.

\subsection{\label{subsec:VATvsRPT}Experimental Assessment of the Difference between VAT and RPT }

\begin{figure}
	\centering
    \begin{subfigure}{0.24\textwidth}
 	\includegraphics[width=\textwidth]{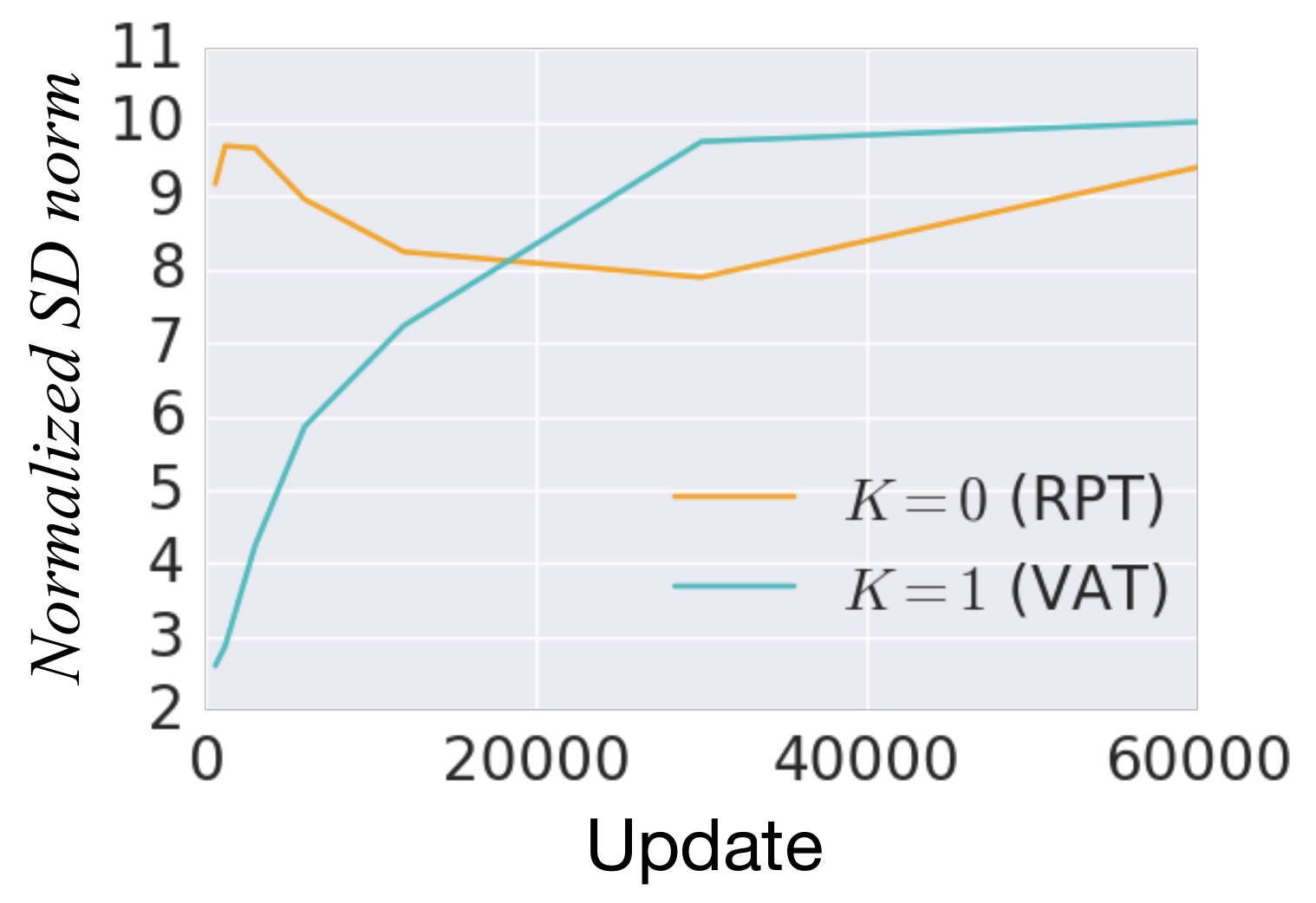}
	\caption{\label{fig:mnist_std_e}MNIST}
    \end{subfigure}
    \begin{subfigure}{0.24\textwidth}
	\includegraphics[width=\textwidth]{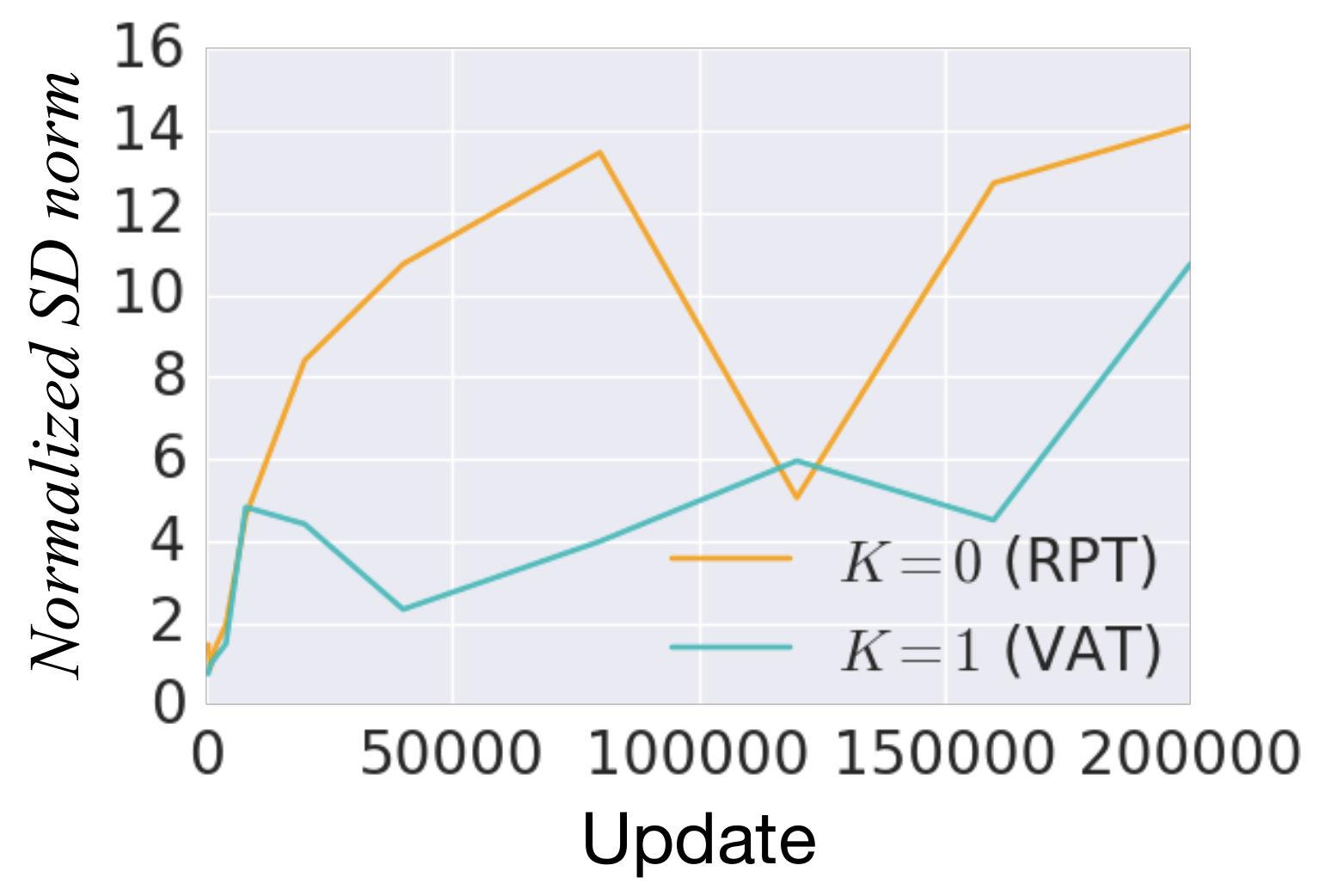}
	\caption{\label{fig:cifar10_std_e}CIFAR-10}
    \end{subfigure}
	\caption{\label{fig:std_e}Transition of the normalized SD norm of $\mathcal{R}^{(0)}$ and $\mathcal{R}^{(1)}$ during VAT training of NNs for supervised learning on MNIST and semi-supervised learning on CIFAR-10. }
\end{figure}

As we showed in the previous section, VAT outperforms RPT in many aspects. There are two possible reasons for the difference in performance. 
First, as mentioned in Section \ref{subsec:vatvsrp}, the power iteration in VAT promotes a \textit{faster learning process} by decreasing the variance of the derivative of the objective function. 
Second, smoothing the function in the direction in which the model is \textit{most sensitive} seems to be much more effective in improving the generalization performance than smoothing the output distribution isotropically around the input.  
We discuss the extent to which these claims might be true.
Let $\mathcal{D}_{M}$ be the mini-batch of size $M$ randomly extracted from $\mathcal{D}$, and let $\hat{\mathcal{R}}^{(K)}(\theta;\mathcal{D}_{M}, r_K)$ be the approximation of $\mathcal{R}^{(K)}(\theta, \mathcal{D})$ computed with the mini-batch $\mathcal{D}_{M}$ and random perturbations $r_K$ generated by $K$-times power iterations. 
To quantify the magnitude of the variance of the stochastic gradient, we define the \textit{normalized standard deviation (SD) norm} for the gradient $\nabla_{\theta}\hat{\mathcal{R}}^{(K)}(\theta;\mathcal{D}_{M},r_K)$ of the regularization term $\hat{\mathcal{R}}^{(K)}(\theta;\mathcal{D}_{M},r_K)$, which is given as the square root of the trace of its variance normalized by the $L_2$ norm of its expectation:
\begin{align}
&{normalized~SD~norm} \nonumber\\
&\equiv \frac{\sqrt{{\rm{trace}}\left(\textrm{Var}_{\mathcal{D}_{M},r_K}\left[\nabla_{\theta}\hat{\mathcal{R}}^{(K)}(\theta;\mathcal{D}_{M},r_{K})\right]\right)}}
{\left\|E_{\mathcal{D}_{M},r_K}\left[\nabla_{\theta}\hat{\mathcal{R}}^{(K)}(\theta;\mathcal{D}_{M},r_{K})\right]\right\|_2},
\end{align}
where $\textrm{Var}_{\mathcal{D}_{M},r_K}[\cdot]$ and $E_{\mathcal{D}_{M},r_K}[\cdot]$ respectively represent the variance and expectation with respect to the randomly selected mini-batch $\mathcal{D}_{M}$ and perturbation $r_K$.
Fig.~\ref{fig:std_e} shows the transition of the normalized SD norm during the VAT process of NNs for the supervised learning task on MNIST (Fig.~\ref{fig:mnist_std_e}) and the semi-supervised learning task on CIFAR-10 (Fig.~\ref{fig:cifar10_std_e}) with $K=0$ and $K=1$ (i.e., RPT and VAT).
We set $M$ to be $100$ and $128$ on MNIST and CIFAR-10 respectively. 
From the figure, we can observe that the normalized SD norm for $K=1$  is lower than that for $K=0$ in most of the early stages of the training for both MNIST and CIFAR-10. Meanwhile, for the instances on which the normalized SD norm of the gradient of RPT falls below that of VAT, the difference is subtle. 

\begin{figure}[ht]
	\centering
    \begin{subfigure}{0.3\textwidth}
	\includegraphics[width=\textwidth]{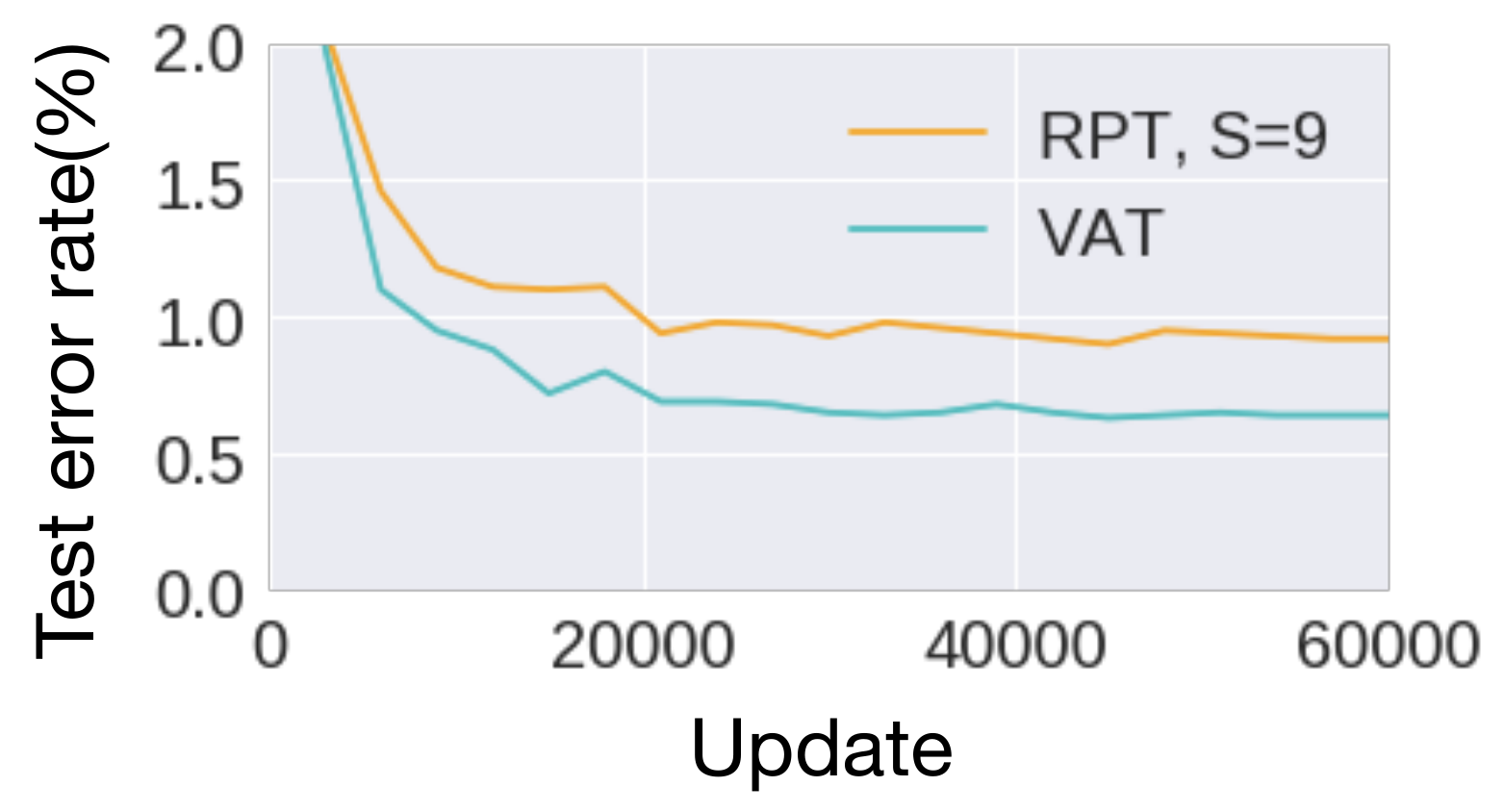}
    \caption{Test error rate}
    \end{subfigure}
    
    \caption{\label{fig:vat_vs_rp}Learning curves of VAT implemented with $\alpha = 1$ and $S=1$ and RPT implemented with optimal $\alpha(=7)$ and $S = 9$.
The hyperparameter $\epsilon$ was set to $2.0$ for both VAT and RPT. 
    }
\end{figure}

For MNIST, the normalized SD norm of RPT becomes as large as $3$ times that of VAT.
To understand how much the normalized SD norm affects the performance, we compared (1) VAT with $\alpha =1$ against (2) RPT implemented with optimal $\alpha$ and an additional procedure of sampling $9=3^2$ random perturbations and using the average of the 9 gradients for each update. Note that the normalized SD norm of the gradient does not depend on $\alpha$.
With this setting, the normalized SD norm of RPT is not greater than that of VAT at the beginning of the training~(Fig.~\ref{fig:vat_vs_rp}).
Remarkably, even with optimal $\alpha$ and increased $S$, the performance of RPT still falls behind that of VAT with $\alpha =1$. 
Even with similar SD norm, the model trained with VAT is more robust against virtual adversarial perturbation than the model trained with RPT.
This, in particular, means that we cannot explain the superiority of VAT over RPT by the reduction in the variance alone.
All these results suggest that the superior performance of VAT owes much to the unique nature of its objective function.
Intuition tells us that the ``max" component in the objective function as opposed to ``expectation" is working in favor of the performance. 
At any phase in the training, the model might lack isotropic smoothness around some sample input points; it is natural that, to fix this, we must endeavor to smooth the model in the direction in which the model is \textit{most} sensitive; that is, we need to take the maximum.

\section{Conclusions}
The results of our experiments on the three benchmark datasets,  MNIST, SVHN, and CIFAR-10 indicate that VAT is an effective method for both supervised and semi-supervised learning. For the MNIST dataset, VAT outperformed recent popular methods other than ladder networks, which are the current state-of-the-art method that uses special network structure. VAT also greatly outperformed the current state-of-the-art semi-supervised learning methods for SVHN and CIFAR-10.  

The simplicity of our method is also worth re-emphasizing. With our approximation of $\mathcal{R}_{\rm vadv}$, VAT can avoid the internal optimization when choosing the adversarial direction and can be implemented with small computational cost. 
Additionally, VAT has only two hyperparameters ($\epsilon$ and $\alpha$) and works sufficiently well on the benchmark dataset with the optimization of $\epsilon$ alone.  
To add even more, VAT is applicable to wide variety of models regardless of its architecture. Also, unlike generative model-based methods, VAT does not require the training of additional models other than the discriminative model~(output distribution) itself.  
At the same time, in our comparative studies, we reconfirmed the effectiveness of generative model-based semi-supervised learning methods on several experimental settings~\cite{kingma2014semi, springenberg2015unsupervised, salimans2016improved}.
Essentially, most generative model-based methods promote generalization performance by making the model robust against perturbations in the region of high $p(x)$ values, the region over which we are likely to receive new input data points in the future. 
In principle, this is complementary to our method, which aims to isotropically smooth the output distribution $p(y|x)$ over each observed input point without any explicit assumption on the input distribution $p(x)$. Combination of these two ideas is a future work.

\section*{Acknowledgments}
This study was supported by the New Energy and Industrial Technology Development Organization (NEDO), Japan.

\bibliography{main}
\bibliographystyle{plain}

\newpage
\onecolumn
\appendices
\section{\label{apd:mnist_exp_set}Supervised Classification for the MNIST Dataset}
The MNIST dataset consists of $28 \times 28$ pixel images of handwritten digits and their corresponding labels. The input dimension is therefore $28 \times 28 = 784$, and each label is one of the numerals from $0$ to $9$.
The following list summarizes the ranges our hyper parameter search:
\begin{tight_itemize}
\item RPT: $\epsilon = [1.0, 50.0]$,
\item adversarial training (with $L_{\infty}$ norm constraint): $\epsilon=[0.05, 0.1]$,
\item adversarial training (with $L_{2}$ norm constraint): $\epsilon=[0.05,5.0]$, and
\item VAT: $\epsilon=[0.05,5.0].$
\end{tight_itemize}
All experiments were conducted with $\alpha = 1$ except when checking the effects of $\alpha$ in Section \ref{subsec:hp}.
Training was conducted using mini-batch SGD based on ADAM~\cite{kingma2014adam}.
We chose the mini-batch size of 100 and used the default values of \cite{kingma2014adam} for the tunable parameters
of ADAM. We trained the NNs with 60,000 parameter updates. For the base learning rate in validation, we selected the initial value of $0.002$, and adopted the schedule of exponential decay with rate $0.9$ per 600 updates. We repeated the experiments 10 times with different random seeds for the weight initialization and reported the mean and standard deviation of the results.

\section{\label{apd:semisup_details}Supervised Classification for CIFAR-10 Dataset}
The CIFAR-10 dataset consists of $32\times 32 \times 3$ pixel RGB images of categorized objects (cars, trucks, planes, animals, and humans). 
The number of training examples and test examples in the dataset are 50,000 and 10,000, respectively. 
We used 10,000 out of 50,000 training examples for validation and we applied ZCA whitening prior to the experiment. 
We also augmented the training dataset by applying random $2 \times 2$ translation and random horizontal flip.
We trained the Conv-Large model~(See Table\ref{tab:cnn_models}) over 300 epochs with batch size 100.
For training, we used ADAM with essentially the same learning rate schedule as the one used in \cite{salimans2016improved}. 
In particular, we set the initial learning rate of ADAM to be $0.003$ and linearly decayed the rate over the last half of training. 
We repeated the experiments 3 times with different random seeds for the weight initialization and reported the mean and standard deviation of the results.

\section{\label{apd:semi-mnist_exp_set}Semi-Supervised Classification for the MNIST Dataset}
For semi-supervised learning of MNIST, we used the same network as the network used for supervised learning; however, we added zero-mean Gaussian random noise with 0.5 standard deviation to the hidden variables during the training. 
This modification stabilized the training on semi-supervised learning with VAT.
We experimented with two sizes of labeled training samples, $N_l\in \{100, 1000\}$, and observed the effect of $N_l$ on the test error. We used the validation set of fixed size(1,000), and used all the training samples, excluding the validation set and labeled training samples, to train the NNs; that is, when $N_l$ = 100, the unlabeled training set $N_{ul}$ had the size of $60,000-100-1,000=58,900$. 

We searched for the best hyperparameter $\epsilon$ from  $[0.05,10.0]$. All experiments were conducted with $\alpha = 1$ and $K =1$. For the optimization method, we again used ADAM-based mini-batch SGD with the same hyperparameter values that we used in supervised setting. We note that the likelihood term can be computed from labeled data only.

We used two separate mini-batches at each step: one mini-batch of size 64 from labeled samples to compute the likelihood term, and another mini-batch of size 256 from both labeled and unlabeled samples to compute the regularization term. 
We trained the NNs over 100,000 parameter updates, and started to decay the learning rate of ADAM linearly after we 50,000-th update. 
We repeated the experiments 3 times with different random seeds for the weight initialization and for the selection of labeled samples. We reported the mean and standard deviation of the results.

\section{\label{apd:semisup_details}Semi-Supervised Classification for the SVHN and CIFAR-10 Datasets}

The SVHN dataset consists of $32\times 32 \times 3$ pixel RGB images of house numbers and their corresponding labels (0--9). The number of training samples and test samples within the dataset are 73,257 and 26,032, respectively. 
We reserved a sample dataset of size 1,000 for validation. From the remainder, we selected sample dataset of size 1,000 as a labeled dataset in semi-supervised training.
Likewise in the supervised learning, we conducted ZCA preprocessing prior to the semi-supervised learning of CIFAR-10. 
We also augmented the training datasets with a random $2 \times 2$ translation.  For CIFAR-10 exclusively, we also applied random horizontal flip as well.
For the labeled dataset, we used 4,000 samples randomly selected from the training dataset, from which we selected 1,000 samples for validation.
We repeated the experiment three times with different choices of labeled and unlabeled datasets on both SVHN and CIFAR-10.

For each benchmark dataset, we decided on the value of the hyperparameter $\epsilon$ based on the validation set. 
We also used a mini-batch of size 32 for the calculation of the negative log-likelihood term and used a mini-batch of size 128 for the calculation of $\mathcal{R}_{\rm vadv}$ in Eq. \eqref{eq:full}. We trained each model with 
48,000 updates. This corresponds to 84 epochs for SVHN and 123 epochs for CIFAR-10. 
We used ADAM for the training. We set the initial learning rate of ADAM to $0.001$ and linearly decayed the rate over the last 16,000 updates. The performance of CNN-Small and CNN-Large that we reported in Section~\ref{subsec:semisup} are all based on the trainings with data augmentation and the choices of $\epsilon$ that we described above.

On SVHN, we tested the performance of the algorithm with and without data augmentation, and used the same setting that we used in the validation experiments for both Conv-Small and Conv-Large. 
For CIFAR-10, however, the models did not seem to converge with 48,000 updates; so we reported the results with 200,000 updates.  
We repeated the experiments 3 times with different random seeds for the weight initialization and for the selection of labeled samples. We reported the mean and standard deviation of the results.

\vskip 0.2in

\begin{table*}
		\centering
		\caption{\label{tab:cnn_models} CNN models used in our experiments on CIFAR-10 and SVHN, based on \cite{springenberg2014striving,salimans2016improved, laine2016temporal}. 
        All the convolutional layers and fully connected layers are followed by batch normalization\cite{ioffe2015batch} except the fully connected layer on CIFAR-10. The slopes of all lReLU\cite{maas2013rectifier} functions in the networks are set to $0.1$.}
		\begin{tabular}{c|c|c}
			\toprule
			Conv-Small on SVHN & Conv-Small on CIFAR-10 & Conv-Large\\
			\midrule
		\multicolumn{3}{c}{32$\times$32 RGB image}  \\
		\midrule	
            3$\times$3 conv. 64 lReLU 		&3$\times$3 conv. 96 lReLU 		& 3$\times$3 conv. 128 lReLU 	\\
            3$\times$3 conv. 64 lReLU 		&3$\times$3 conv. 96 lReLU 		& 3$\times$3 conv. 128 lReLU 	\\
            3$\times$3 conv. 64 lReLU		&3$\times$3 conv. 96 lReLU		& 3$\times$3 conv. 128 lReLU	\\
		\midrule
		\multicolumn{3}{c}{2$\times2$ max-pool, stride 2}  \\
		\multicolumn{3}{c}{dropout, $p=0.5$}  \\
            \midrule   
            3$\times$3 conv. 128 lReLU 		&3$\times$3 conv. 192 lReLU 		& 3$\times$3 conv. 256 lReLU 	\\
            3$\times$3 conv. 128 lReLU 		&3$\times$3 conv. 192 lReLU 		& 3$\times$3 conv. 256 lReLU 	\\
            3$\times$3 conv. 128 lReLU 		&3$\times$3 conv. 192 lReLU 		& 3$\times$3 conv. 256 lReLU 	\\
            \midrule  
		\multicolumn{3}{c}{2$\times2$ max-pool, stride 2}  \\
		\multicolumn{3}{c}{dropout, $p=0.5$}  \\
            \midrule 
            3$\times$3 conv. 128 lReLU 		&3$\times$3 conv. 192 lReLU 		& 3$\times$3 conv. 512 lReLU 	\\
            1$\times$1 conv. 128 lReLU 		&1$\times$1 conv. 192 lReLU 		& 1$\times$1 conv. 256 lReLU 	\\
            1$\times$1 conv. 128 lReLU 		&1$\times$1 conv. 192 lReLU 		& 1$\times$1 conv. 128 lReLU 	\\
                \midrule
		\multicolumn{3}{c}{global average pool, 6$\times$6 $\rightarrow$ 1$\times$1 }  \\
            \midrule   
            dense 128 $\rightarrow$ 10	 		&dense 192$\rightarrow$ 10 	&dense 128$\rightarrow$ 10 \\
            \midrule
		\multicolumn{3}{c}{10-way softmax}  \\
		\bottomrule
		\end{tabular}
\end{table*}

\end{document}